\documentclass[runningheads]{llncs}
\usepackage{graphicx}
\usepackage{amsmath,amssymb} 
\usepackage{color}
\usepackage{epsfig}
\usepackage{booktabs}
\usepackage{subcaption}
\usepackage{multirow}
\usepackage{array}
\usepackage{colortbl}
\usepackage{url}

\newcommand{\etal}{\textit{et al}. }
\newcommand{\ie}{\textit{i}.\textit{e}. }
\newcommand{\eg}{\textit{e}.\textit{g}., }

\newcommand{\refFig}[1]{Fig.~\ref{#1}}
\newcommand{\refSec}[1]{Sec.~\ref{#1}}
\newcommand{\refEq}[1]{Equation~(\ref{#1})}
\newcommand{\refTab}[1]{Table~\ref{#1}}

\definecolor{lightgray}{gray}{0.8}

\begin{document}

\title{Deep Depth From Focus\thanks{This research was partially funded by the Humboldt Foundation through the Sofja Kovalevskaja Award and ERC Consolidator Grant ``3D Reloaded''.}} 
\titlerunning{Deep Depth From Focus} 


\author{Caner Hazirbas \and Sebastian Georg Soyer \and Maximilian Christian Staab\and\\ Laura Leal-Taix\'e \and Daniel Cremers}
%
\index{Soyer, Sebastian Georg}
\index{Staab, Maximilian Christian}

\authorrunning{C. Hazirbas et al.} 


\institute{Technical University of Munich, Germany\\
\email{\{hazirbas, soyers, staab, leal.taixe, cremers\}@cs.tum.edu}}

\maketitle

\begin{abstract}
	Depth from focus (DFF) is one of the classical ill-posed inverse problems in computer vision. Most approaches recover the depth at each pixel based on the focal setting which exhibits maximal sharpness. Yet, it is not obvious how to reliably estimate the sharpness level, particularly in low-textured areas.  In this paper, we propose `Deep Depth From Focus (DDFF)' as the first end-to-end learning approach to this problem. One of the main challenges we face is the hunger for data of deep neural networks. In order to obtain a significant amount of focal stacks with corresponding groundtruth depth, we propose to leverage a light-field camera with a co-calibrated RGB-D sensor. This allows us to digitally create focal stacks of varying sizes. Compared to existing benchmarks our dataset is 25 times larger, enabling the use of machine learning for this inverse problem. We compare our results with state-of-the-art DFF methods and we also analyze the effect of several key deep architectural components.  These experiments show that our proposed method `DDFFNet' achieves state-of-the-art performance in all scenes, reducing depth error by more than 75\% compared to the classical DFF methods.
\keywords{depth from focus \and convolutional neural networks}
\end{abstract}
\section{Introduction}
The goal of \textit{depth from focus} (DFF) is to reconstruct a
pixel-accurate disparity map given a stack of images with
gradually changing optical focus.  The key observation is that a
pixel's sharpness is maximal when the object it belongs to is in
focus. Hence, most methods determine the depth at each pixel by
finding the focal distance at which the contrast measure is maximal.
Nonetheless, DFF is an ill-posed problem, since this assumption does
not hold for all cases, especially for textureless surfaces where
sharpness cannot be determined.  This is why most methods rely on
strong regularization to obtain meaningful depth maps which in turn
leads to an often oversmoothed output.

\begin{figure}
	\newcommand\scale{0.115}
	\centering
	\begin{tabular}{cccccccc}
		Image & Disparity & VDFF & DDLF & PSP-LF & Lytro & PSPNet & Proposed
		\\ \toprule
		\includegraphics[width=\scale\textwidth,trim={0cm 4cm 5cm 0}, clip]{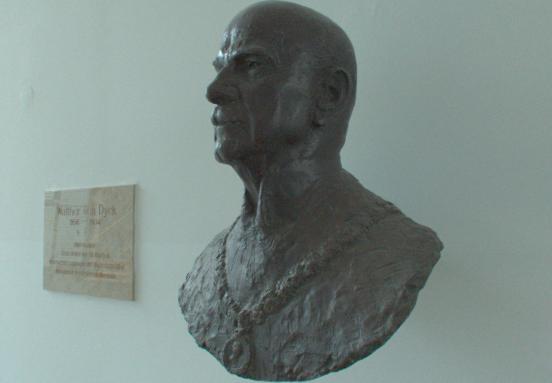}&
		\includegraphics[width=\scale\textwidth,trim={0cm 4cm 5cm 0}, clip]{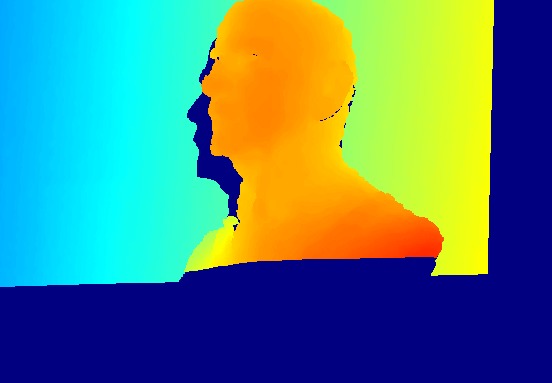}&
		\includegraphics[width=\scale\textwidth,trim={0cm 4cm 5cm 0}, clip]{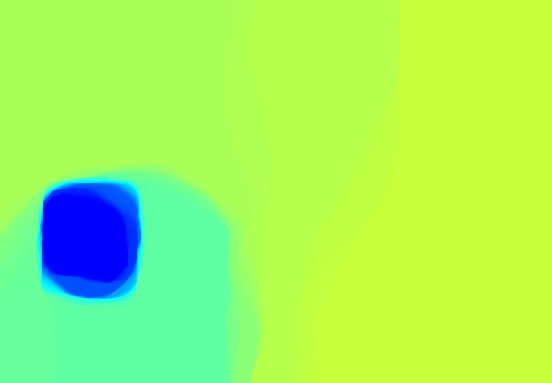}&
		\includegraphics[width=\scale\textwidth,trim={0cm 4cm 5cm 0}, clip]{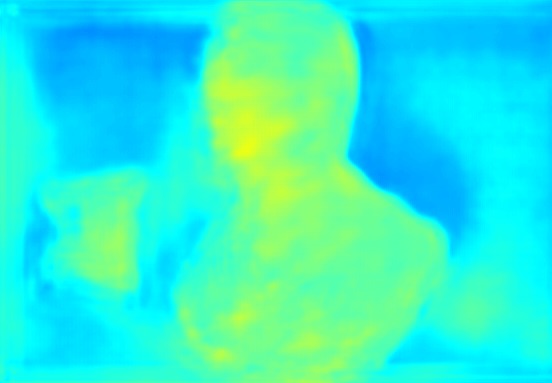}&
		\includegraphics[width=\scale\textwidth,trim={0cm 4cm 5cm 0}, clip]{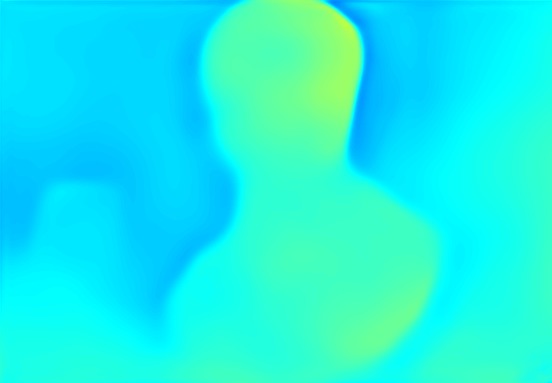}&
		\includegraphics[width=\scale\textwidth,trim={0cm 4cm 5cm 0}, clip]{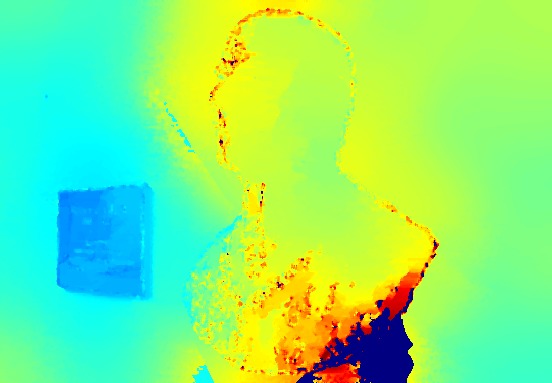}&
		\includegraphics[width=\scale\textwidth,trim={0cm 4cm 5cm 0}, clip]{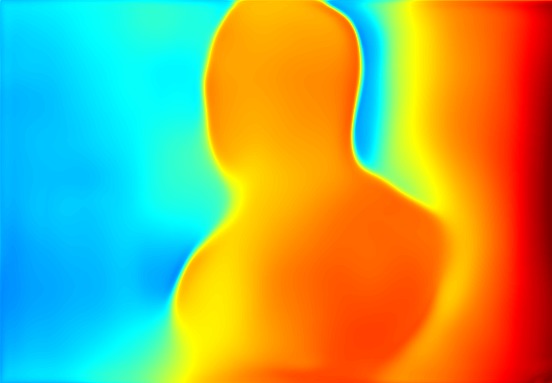}&
		\includegraphics[width=\scale\textwidth,trim={0cm 4cm 5cm 0}, clip]{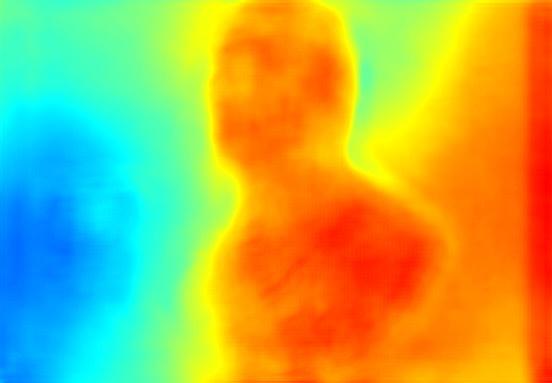}\\
		
		\includegraphics[width=\scale\textwidth,trim={0cm 4cm 5cm 0}, clip]{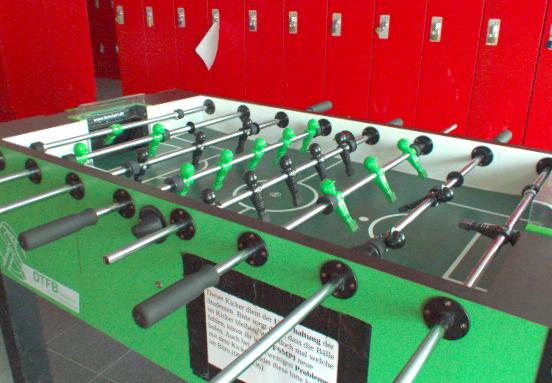}&
		\includegraphics[width=\scale\textwidth,trim={0cm 4cm 5cm 0}, clip]{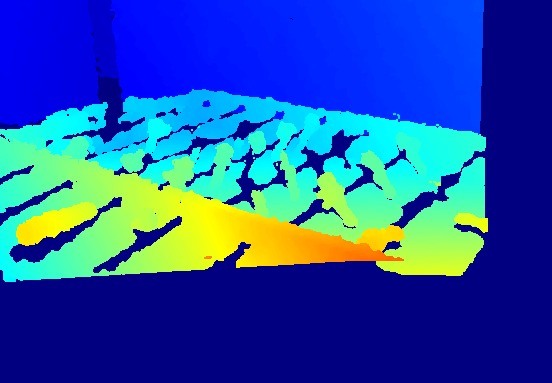}&
		\includegraphics[width=\scale\textwidth,trim={0cm 4cm 5cm 0}, clip]{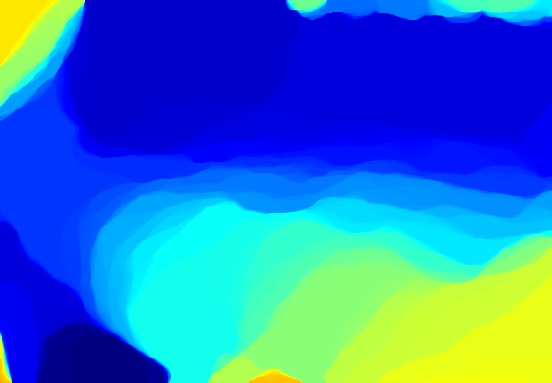}&
		\includegraphics[width=\scale\textwidth,trim={0cm 4cm 5cm 0}, clip]{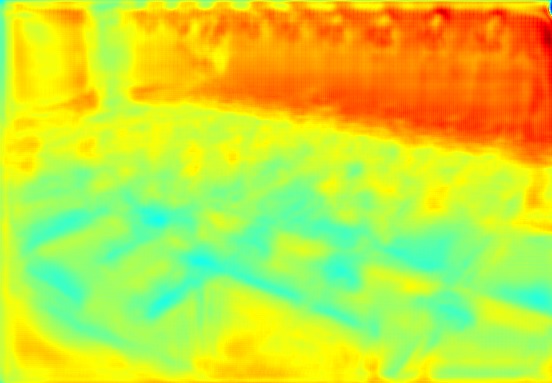}&
		\includegraphics[width=\scale\textwidth,trim={0cm 4cm 5cm 0}, clip]{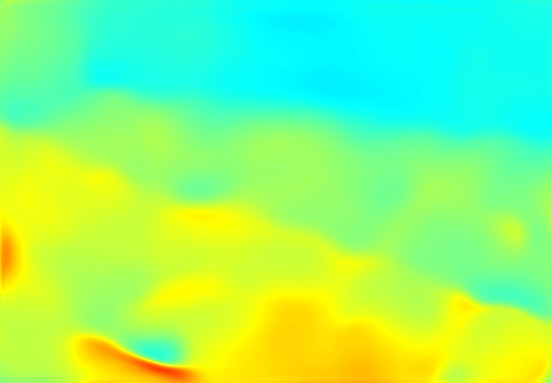}&
		\includegraphics[width=\scale\textwidth,trim={0cm 4cm 5cm 0}, clip]{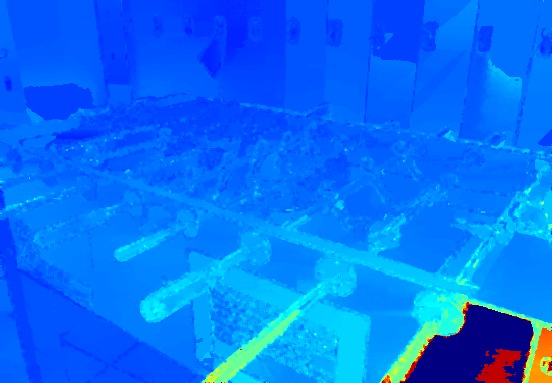}&
		\includegraphics[width=\scale\textwidth,trim={0cm 4cm 5cm 0}, clip]{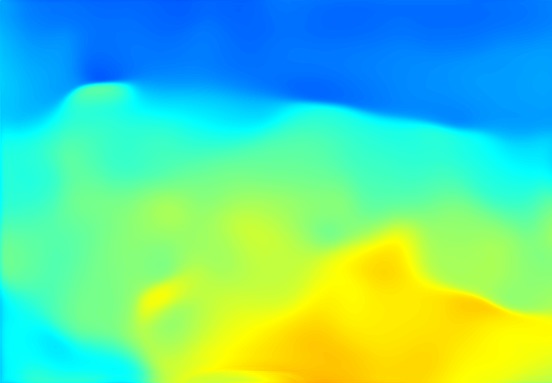}&
		\includegraphics[width=\scale\textwidth,trim={0cm 4cm 5cm 0}, clip]{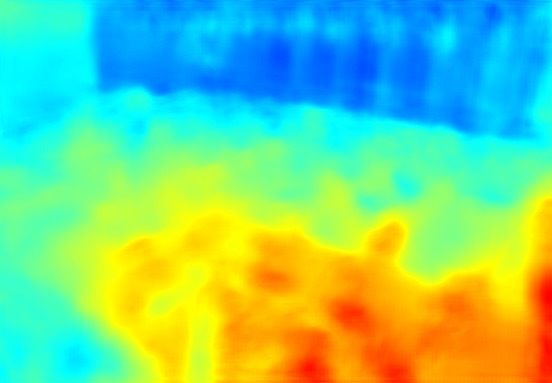}\\	
	\end{tabular}
	\caption{\small \textbf{Qualitative results of the DDFFNet versus state-of-the-art methods.} Results are normalized by the maximum disparity, which the focal stacks are generated on (0.28 pixel). Warmer colors represent closer distances. Best viewed in color.}
	\label{fig:qualRes}
\end{figure}

While spatial smoothness is a rather primitive prior for depth reconstruction, with the advent of Convolutional Neural Networks (CNNs) we now have an alternative technique to resolve classical ill-posed problems such as semantic segmentation \cite{badrinarayanan2017segnet,hazirbasma16fusenet,long15fcn,noh15deconvnet} or optical flow estimation \cite{dosovitsky15flownet}. The underlying expectation is that the rather naive and generic spatial smoothness assumption used in variational inference techniques is replaced with a more object-specific prior knowledge absorbed through huge amounts of training data.

A big strength of CNNs is their ability to extract meaningful image features, and correlate pixel information via convolutions. Our intuition is that a network will be able to find the image in the focal stack at which a pixel is maximally sharp, thereby correlating focus and depth.
We therefore propose to tackle the task of depth from focus using end-to-end-learning. 
To that end, we create the first DFF dataset with real-world scenes and groundtruth measured depth. In order to obtain focal stacks in a reliable and fast way, we propose to use a light-field camera. Also called plenoptic camera, it allows us to obtain multi-view images of a scene with a single photographic exposure. All-in-focus images as well as focal stacks can be recovered digitally from a light-field image. 
Using this new dataset, we perform end-to-end learning of the disparity given a focal stack.

\subsection{Contribution}

We present \textit{Deep Depth From Focus Network} (DDFFNet), an auto-encoder-style Convolutional Neural Network that outputs a disparity map from a focal stack.
To train such a net, we create a dataset with 720 light-field images captured using a plenoptic camera,~\ie Lytro ILLUM, covering 12 indoor scenes. Given a light-field image, we can digitally generate a focal stack.
Groundtruth depth is obtained from an RGB-D sensor which is calibrated to the light-field camera.

To the best of our knowledge, this is the largest dataset with groundtruth for the problem of DFF.
We experimentally show that this amount of data is enough to successfully fine-tune a network. We compare our results with state-of-the-art DFF methods and provide a comprehensive study on the impact of different variations of the encoder-decoder type of network.

\noindent The contribution of this paper is three-fold:
\begin{itemize}
	\item We propose DDFFNet, the first end-to-end learning method to compute depth maps from focal stacks.
	\item We introduce DDFF 12-Scene: a dataset with 720 light-field images and co-registered groundtruth depth maps recorded with an RGB-D sensor. We show that this data is enough to train a network for the task of DFF.
	\item We compare several state-of-the-art methods for DFF, as well as several variations of the encoder-decoder architecture, and show that our method outperforms the other methods by a large margin.  It computes depth maps in 0.6 seconds on an NVidia Pascal Titan X GPU.
\end{itemize}

\subsection{Related work}
\noindent{\bf Depth from focus or Shape from focus.}
Conventional methods aim at determining the depth of a pixel by measuring its sharpness or focus at different images of the focal stack~\cite{pertuz13analysis}. Developing a discriminative measure for sharpness is non trivial, we refer the reader to~\cite{pertuz13analysis} for an overview. 
Other works aim at filtering the contrast coefficients before determining depth values by windowed averaging~\cite{thelentip2009} or non-linear filtering~\cite{mahmoodtip2012}. 
Another popular approach to obtain consistent results is to use total variation regularization. 
~\cite{mahmood13tv} proposed the first variational approach to tackle DFF while 
~\cite{moeller15vdff} defines an objective function composed of a smooth but nonconvex data term with a non-smooth but convex regularizer to obtain a robust (noise-free) depth map.
Suwajanakorn~\etal~\cite{suwajanakorn15ddfmobile} computes DFF on mobile devices, focusing on compensating the motion between images of the focal stack. 
This results in a very involved model, that depends on optical flow results, and takes 20 minutes to obtain a depth map.
Aforementioned methods heavily rely on priors/regularizers to increase the robustness of the algorithm, meaning their models may not generalize to all scenes. Interestingly, shape from focus was already tackled using neural networks in 1999~\cite{asifspie1999}, showing their potential on synthetic experiments. The increasing power of deep architectures 
makes it now possible to move towards estimating depth of real-world scenarios.\\

\noindent{\bf Plenoptic or light-field cameras.}
A light-field or plenoptic camera captures angular and spatial information on the distribution of light rays in space. In a single photographic exposure, these cameras are able to obtain multi-view images
of a scene. The concept was first proposed in \cite{plenoptic}, and has recently gained interest from the computer vision community. 
These cameras have evolved from bulky devices~\cite{bulkycamera} to hand-held cameras based on micro-lens arrays~\cite{microlenscamera}. 
Several works focus on the calibration of these devices, either by using raw images and line features~\cite{bokpami2017} or by decoding 2D lenslet images into 4D light-fields~\cite{dansereaucvpr2013}. 
An analysis of the calibration pipeline is detailed in ~\cite{choiccv2013}.
Light-field cameras are particularly interesting since depth and all-in-focus images can be computed directly from the 4D light-field~\cite{jeoncvpr2015,liniccv2015,perez3dtv2009}. 
Furthermore, focal stacks, \ie images
taken at different optical focuses, can be obtained from plenoptic cameras with a single photographic exposure.
For this reason, we choose to capture our training dataset using these cameras, though any normal camera that captures images at different optical focuses can be used at test time.

To the best of our knowledge, there are only two light-field datasets with groundtruth depth maps~\cite{honauer16benchmark,wanner13datasets}.  
While \cite{wanner13datasets} provides 7 synthetic and only 6 real-scene light-fields, ~\cite{honauer16benchmark} 
generates a hand-crafted synthetic light-field benchmark composed of only 24 samples with groundtruth disparity maps.
Our dataset is 25 times larger, composed of 12 indoor scenes, in total of 720 light-field samples with co-registered groundtruth depth obtained from an RGB-sensor, ranging from 0.5 to 7$m$.
In this work, we show that our data is enough to fine-tune a network for the specific task of predicting depth from focus.

Note that the Stanford Light-field dataset\footnote{\url{http://lightfields.stanford.edu/}} has more samples than our dataset, but does not provide {\it groundtruth depth maps}. The depth maps in this dataset were produced by the standard Lytro toolbox which outputs lambda-scaled depth maps. Therefore, the maps are not in real distance metrics and there is no camera calibration provided. Furthermore, Lytro depth maps can be inaccurate as we will show in~\refSec{sec:experiments}, and cannot be considered as groundtruth.\\

\noindent{\bf Deep learning.}
Deep learning has had a large impact in computer vision since showing its excellent performance in the task of image classification~\cite{he16resnet,krizhevsky12alexnet,simonyan15vgg}. A big part of its success has been the creation of very large annotated datasets such as ImageNet~\cite{russakovsky15imagenet}. Of course, this can also be seen as a disadvantage, since creating such datasets with millions of annotations for each task would be impractical. Numerous recent works have shown that networks pre-trained on large datasets for seemingly unrelated tasks like image classification, can easily be fine-tuned to a new task for which there exists only a fairly small training dataset. This paradigm has been successfully applied to object detection~\cite{girshick15fastrcnn}, pixel-wise semantic segmentation~\cite{badrinarayanan2017segnet,hazirbasma16fusenet,kendall15bayesiansegnet,long15fcn,noh15deconvnet,zhao17pspnet}, depth and normal estimation~\cite{gargeccv2016,licvpr2015,liu16learningdepth} or single image-based 3D localization~\cite{kendall2015posenet,Walch2017ICCV}, to name a few. Another alternative is to generate synthetic data to train very large networks,~\eg for optical flow estimation~\cite{dosovitsky15flownet}. Using synthetic data for training is not guaranteed to work, since the training data often does not capture the real challenge and noise distribution of real data. Several works use external sources of information to produce groundtruth. \cite{gallianicvpr2016} uses sparse multi-view reconstruction results to train a CNN to predict surface normals, which are in turn used to improve the reconstruction. In~\cite{gargeccv2016}, the authors aim at predicting depth from a single image, but create groundtruth depth data from matching stereo images. We propose to use an RGB-D sensor that can be registered to our light-field camera to obtain the groundtruth depth map. Even though an RGB-D sensor is not noise-free, we show that the network can properly learn to predict depth from focus even from imperfect data. We use the paradigm of fine-tuning a pre-trained network and show that this works even if the tasks of image classification and DFF seem relatively unrelated.

\section{DDFF 12-Scene benchmark}
\label{sec:sixscene}
In this section we present our indoor DDFF 12-Scene dataset for \textit{depth from focus}. 
This dataset is used for the training and evaluation of the proposed and several state-of-the-art methods.
We first give the details on how we generate our data, namely the focal stack and groundtruth depth maps.\\

\noindent{\bf Why a 4D light-field dataset for depth from focus?} To determine the depth of a scene from focus, we first need to generate a focal stack obtainable by using any camera and changing the focal step manually to retrieve the refocused images. Nonetheless, this is a time-consuming task that would not allow us to collect a significant amount of data as it is required to train deep models. Given the time that it takes to change the focus on a camera, the illumination of the scene could have easily changed or several objects could have moved.
We therefore propose to leverage a light-field camera as it has the following advantages: (i) only one image per scene needs to be taken, meaning all images will have the same photographic exposure and the capturing process will be efficient, (ii) refocusing can be performed digitally, which allows us to easily generate stacks with different focal steps,
(iii) the dataset can be a benchmark not only for DFF, but also for other tasks such as depth from light-field or 3d reconstruction from light-field.
Even though we do not intend to tackle these tasks in this work, we do show some comparative results on depth from light-Field in~\refSec{sec:experiments}.\\

\noindent{\bf Light-field imaging.} With light-field imaging technology, the original focus of the camera can be altered after the image is taken. Following this, we use a commercially available light-field camera,~\ie Lytro ILLUM\footnote{{Lytro ILLUM} lightfield camera, \url{illum.lytro.com}, accessed: 2016-11-07},  
to collect data and then generate focal stacks. Plenoptic cameras capture a 4D light-field $L(u,v,x,y)$ which stores the light rays that intersect the image plane $\Omega$ at $(x,y)$ and the focus or camera plane $\Pi$ at $(u,v)$. The pixel intensity $I(x,y)$ is then:
\begin{equation}
I(x,y) = \int_u \int_v L(u,v,x,y)\,\,\, \partial u\, \partial v\,.
\end{equation}
Refocusing on an image corresponds to shifting and summing all sub-apertures, $I_{(u,v)}(x,y)$. Given the amount of shift, pixel intensities of a refocused image are computed as follows~\cite{diebold13refocus}:
\begin{equation}
I'(x,y) = \int_u \int_v L(u,v,x+\Delta_x(u),y+\Delta_y(v))\, \partial u\, \partial v\,.
\end{equation}
The shift ($\Delta_{u}$, $\Delta_{v}$) of each sub-aperture $uv$ can be physically determined given an arbitrary depth $Z$ in $m$, at which the camera is in-focus:
\begin{equation}
\begin{pmatrix} \Delta_{x}(u) \\ \Delta_{y}(v) \end{pmatrix}  = \underbrace{\frac{\text{baseline} \cdot f}{Z}}_\text{disparity}\cdot \begin{pmatrix} u_{center} - u \\ v_{center} -v \end{pmatrix} \, ,
\label{eq:refocus}
\end{equation}
where the baseline is the distance between adjacent sub-apertures in meter/pixel, $f$ is the focal length of the microlenses in pixels and ($u,\,v$)$^T$ indicates the spatial position of the sub-aperture in the $\Pi$ plane in pixels. Although shifting can be performed using bilinear or bicubic interpolation, to be able to perform subpixel accurate focusing on the images, following~\cite{jeoncvpr2015} we use the \textit{phase shift algorithm} to observe the impact of subpixel shifts on the images:
\begin{equation}
\mathcal{F}\{I'(x+\Delta_x(u))\} = \mathcal{F}\{I(x)\} \cdot \exp^{2\pi i \Delta_x(u)} \,\, ,
\end{equation}
where $\mathcal{F}\{\cdot\}$ is the 2D discrete Fourier transform.
We generate the focal stacks within a given disparity range, for which the focus shift on the images is clearly observable from close objects to far ones present in our dataset. Disparity values used in refocusing in~\refEq{eq:refocus} are sampled linearly in the given interval for a stack size of $S$, meaning that the focus plane equally shifts in-between the refocused images. Example refocused images for disparity $\in\{0.28, 0.17, 0.02\}$ are shown in~\refFig{fig:refocus}. Note that we chose to use a light-field camera since it is easy to obtain a focal stack from it. Nonetheless, at test time, any imaging device could be used to take images at different optical focus.\\

\begin{figure}[t!]
	\centering
	\begin{subfigure}[t]{0.46\textwidth}
		\includegraphics[width=\textwidth]{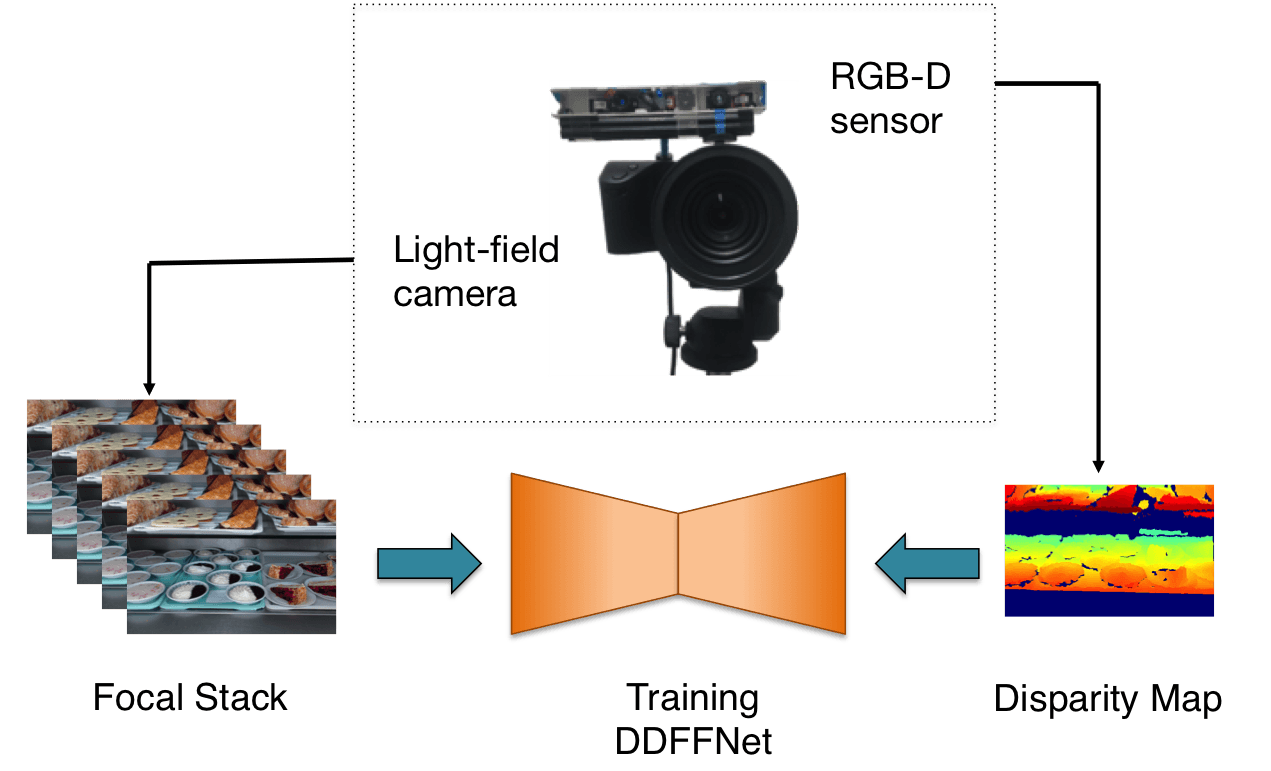}
		\caption{\label{fig:camera_setup}}
	\end{subfigure}
	\qquad
	\begin{subfigure}[t]{0.44\textwidth}
		\includegraphics[width=\textwidth]{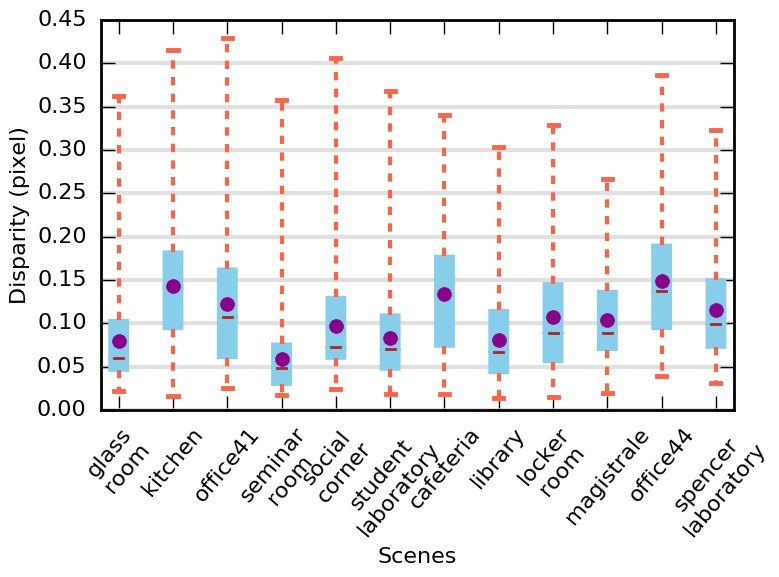}
		\caption{\label{fig:dispwhisker}}
	\end{subfigure}
	\caption{\small\textbf{(a) Experimental setup}: we place an RGB-D sensor on top of a plenoptic camera in order to capture calibrated groundtruth depth maps and light-field images from which we then create focal stacks. These two inputs will be used to train the \textit{DDFFNet}. \textbf{(b) Whisker diagram} of the disparity distribution for each scene. Circle and red line are the mean and median, respectively. Minimum disparity is 0.015 pixels (bottom orange lines), maximum disparity is 0.43 pixels (top orange lines).}
\end{figure}

\noindent{\bf Light-field camera calibration.} For consistent capturing over all scenes, we fix the focal length of the main lens to $f=9.5mm$ and lock the zoom.
To increase the re-focusable range of the camera, we use the \textit{hyperfocal mode}. Theoretically, we can then refocus from $27cm$ distance to infinity. We set the white-balancing, ISO and shutter speed settings to \textit{auto} mode. In order to estimate the intrinsic parameters of the light-field camera, we use the calibration toolbox by Bok~\etal~\cite{bokpami2017} with a chessboard  pattern composed of $26.25mm$ length squares. The radius of each microlens is set to 7 pixels as suggested in~\cite{bokpami2017}.

We can generate $9\times9$ undistorted sub-apertures, each of which has $383\times552$ image resolution. Estimated intrinsic parameters of the microlenses are: focal length $f=f_x=f_y=521.4$, microlens optical center $(c_x,\, c_y)^T = (285.11,\, 187.83)^T$ and the baseline (distance between two adjacent sub-apertures in meter/pixel) is $27\mathrm{e}{-5}$. All estimated parameters can be found in the supplementary material.\\

\noindent{\bf Groundtruth depth maps from an RGB-D sensor.} Along with the light-field images, we also provide groundtruth depth maps. 
To this end, we use an RGB-D structure sensor,~\ie ASUS Xtion PRO LIVE, and mount it on the hot shoe of the light-field camera (see~\refFig{fig:camera_setup}). 
Since we only need the infrared camera of the RGB-D sensor, we align the main lens of Lytro ILLUM to the infrared image sensor as close as possible for a larger overlap on the field of views of both cameras. 
We save the $480\times640$ resolution depth maps in millimeters. 
RGB-D sensors are not accurate on glossy surfaces and might even produce a large amount of invalid/missing measurements. 
In order to reduce the number of missing values, we take nine consecutive frames and save the median depth of each pixel  during recording/capturing.\\

\begin{figure}[!t]
	\centering
	\begin{tabular}{ccccc}
		\includegraphics[width=0.18\textwidth,trim={0cm 4cm 3cm 0},clip]{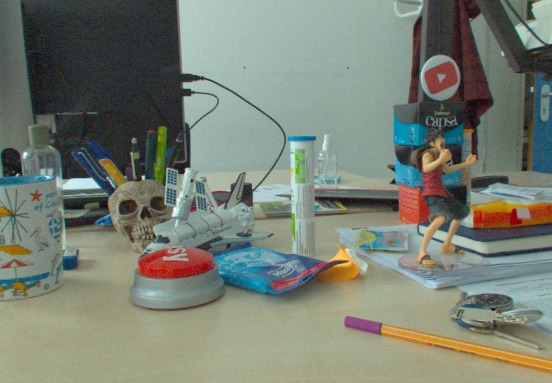}&
		\includegraphics[width=0.18\textwidth,trim={0cm 4cm 3cm 0},clip]{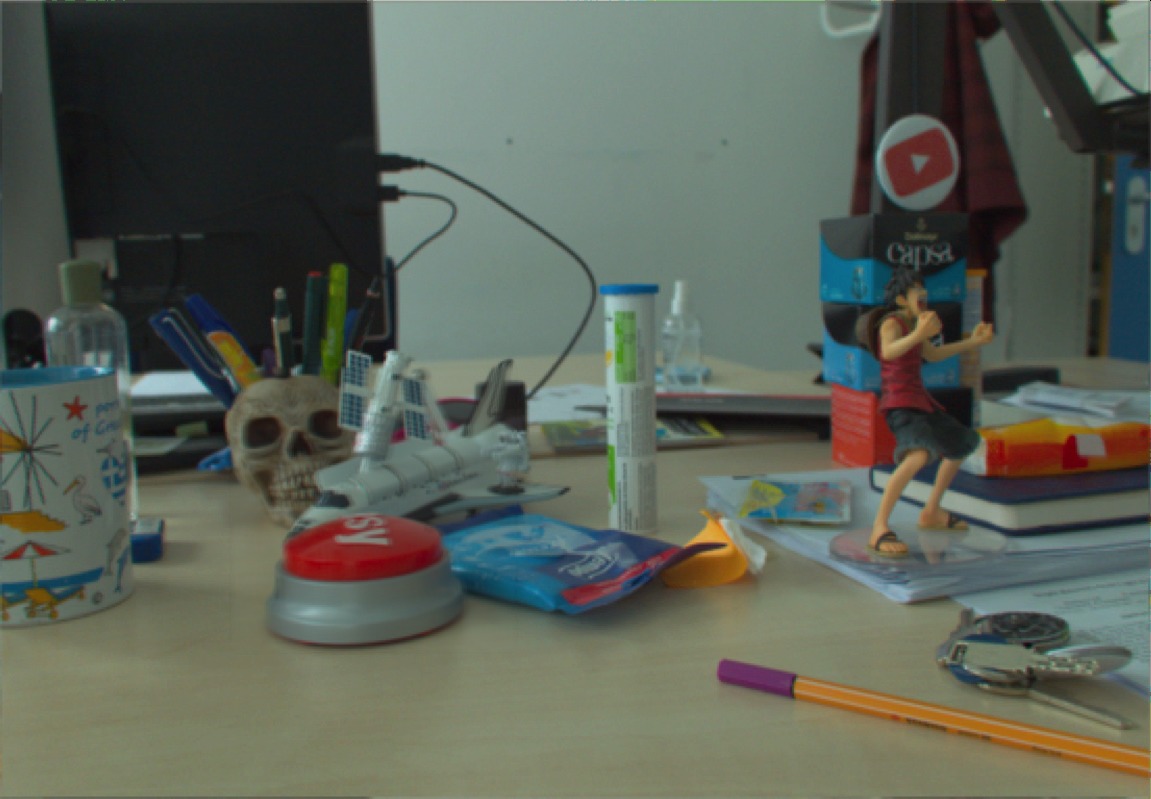}&
		\includegraphics[width=0.18\textwidth,trim={0cm 4cm 3cm 0},clip]{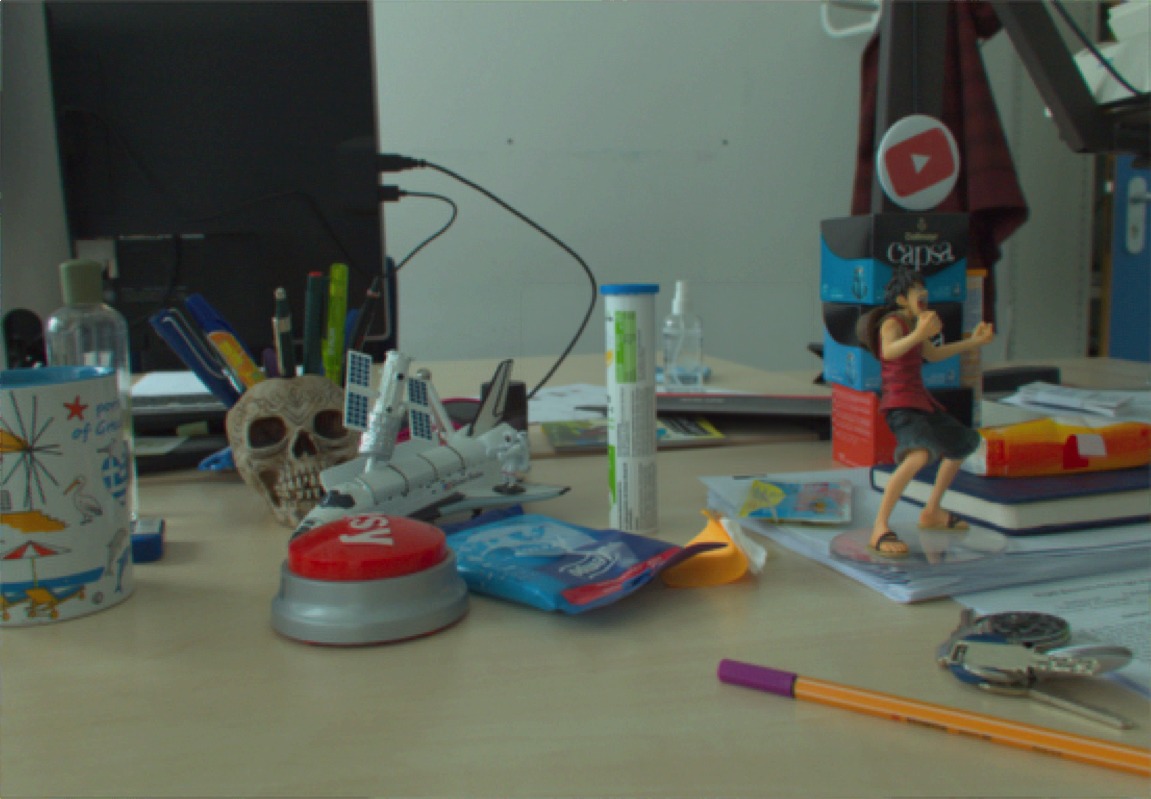}&
		\includegraphics[width=0.18\textwidth,trim={0cm 4cm 3cm 0},clip]{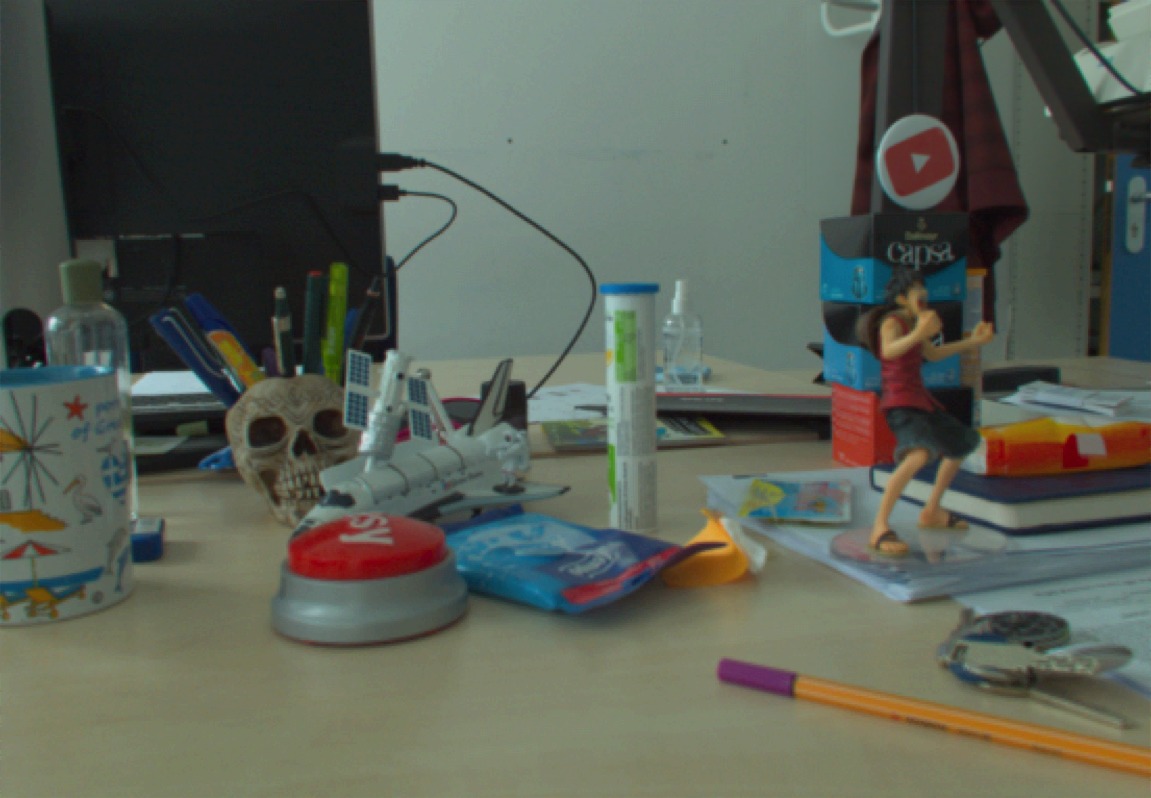}&
		\includegraphics[width=0.18\textwidth,trim={0cm 4cm 3cm 0},clip]{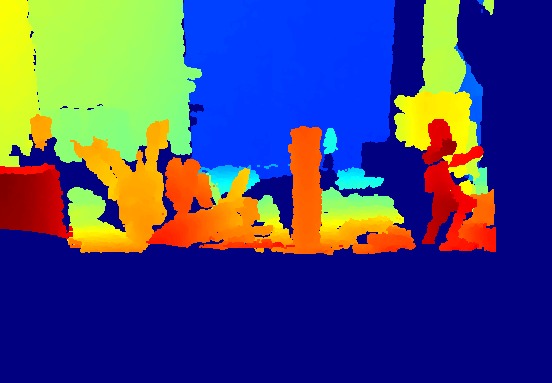}\\
		
		\includegraphics[width=0.18\textwidth,trim={0cm 4cm 3cm 0},clip]{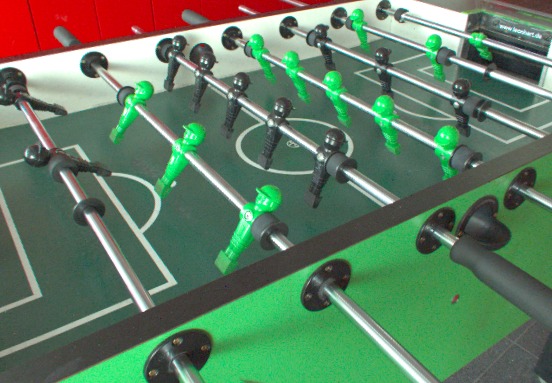}&
		\includegraphics[width=0.18\textwidth,trim={0cm 4cm 3cm 0},clip]{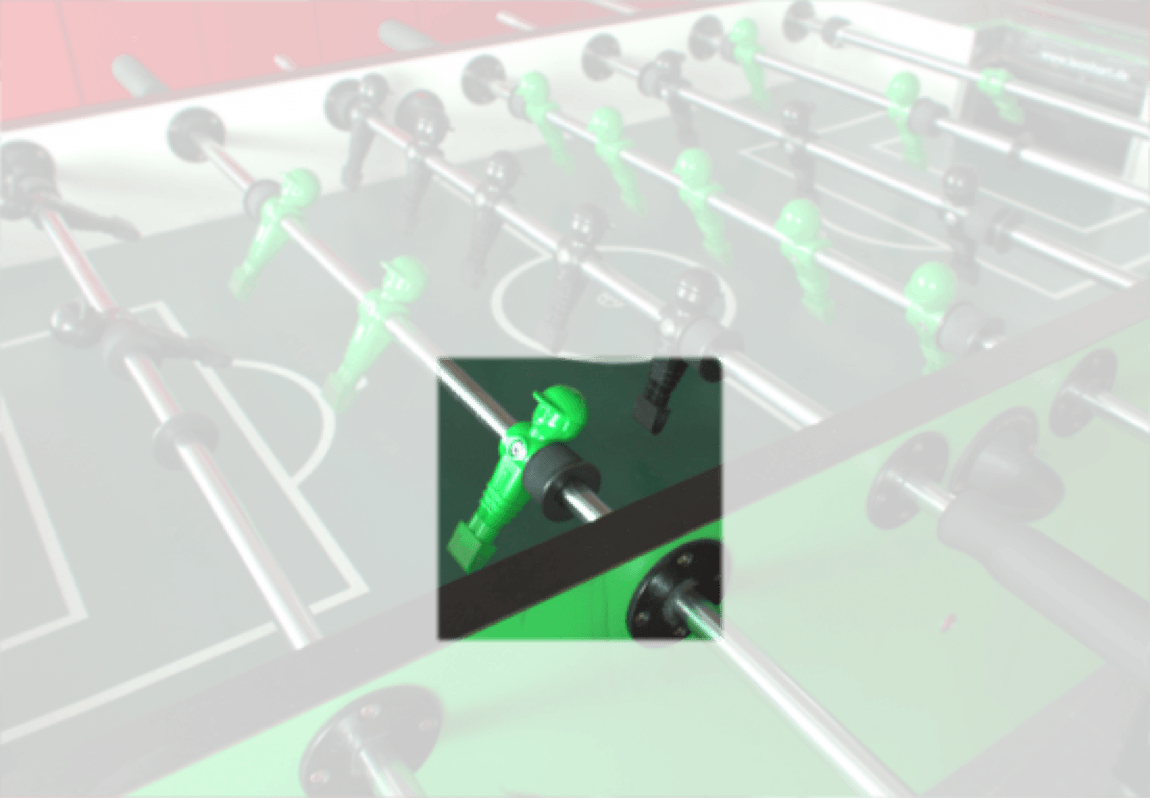}&
		\includegraphics[width=0.18\textwidth,trim={0cm 4cm 3cm 0},clip]{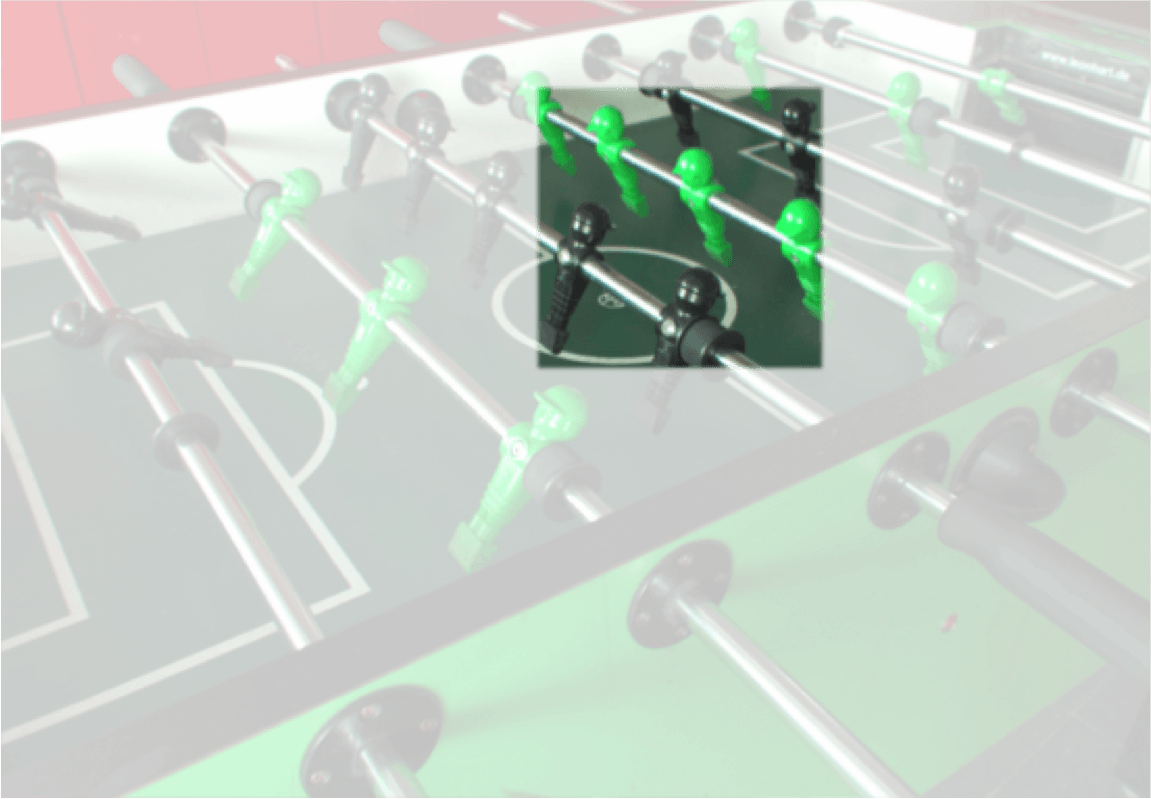}&
		\includegraphics[width=0.18\textwidth,trim={0cm 4cm 3cm 0},clip]{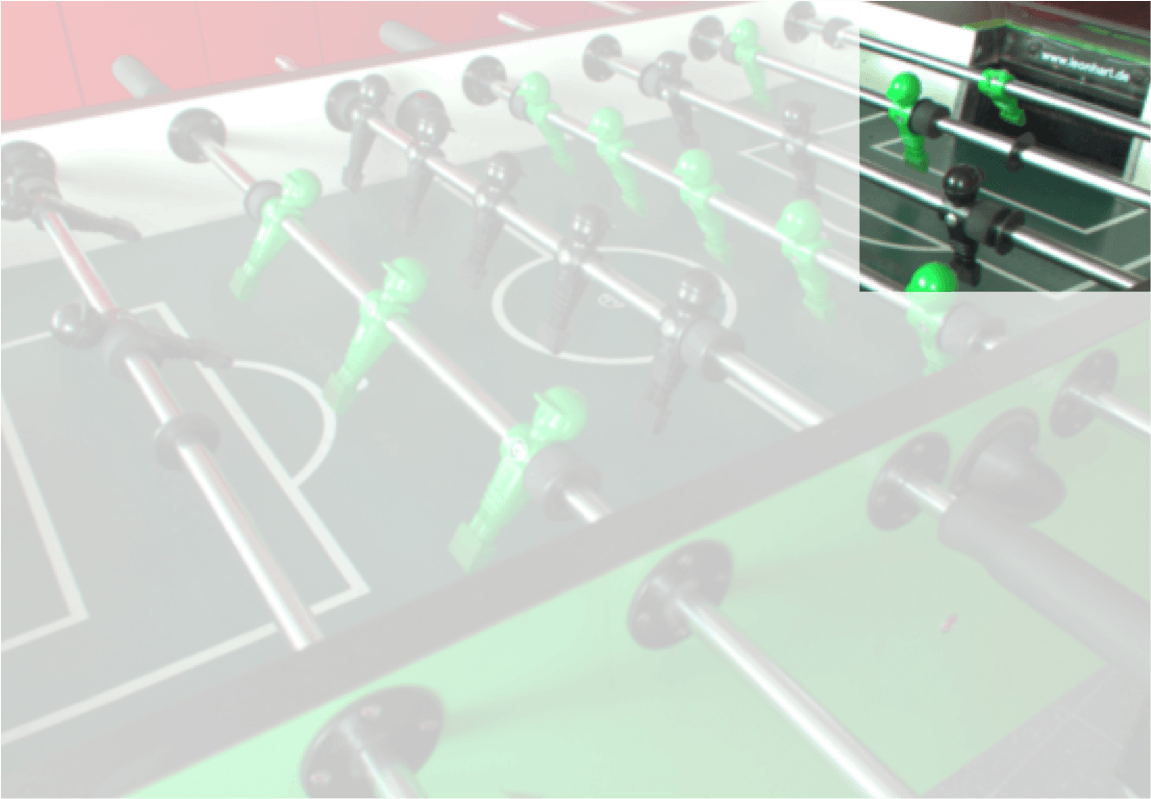}&
		\includegraphics[width=0.18\textwidth,trim={0cm 4cm 3cm 0},clip]{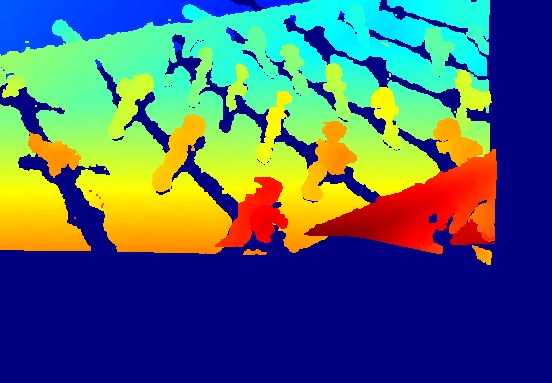}\\
		\scriptsize Center sub-aperture & 
		\scriptsize disparity$=0.28$ &
		\scriptsize disparity$=0.17$ & 
		\scriptsize disparity$=0.02$ & 
		\scriptsize GT Disparity \\
	\end{tabular}
	\caption{\small \textbf{Example refocused images.} {\it First column:} Center sub-aperture image. {\it Last column:} groundtruth disparity maps from the RGB-D sensor. {\it Middle columns:} Refocused images for varying disparity values in pixels, regions in focus are highlighted. Best viewed in color.}
	\label{fig:refocus}
\end{figure}

\noindent{\bf Stereo camera calibration.} We perform mono and stereo camera calibration to estimate the relative pose of the depth sensor with respect to the light-field camera.
To this end, we use the publicly available \textit{Camera Calibration Toolbox for Matlab}\footnote{\url{www.vision.caltech.edu/bouguetj/calib_doc/}}. 
We use the same calibration pattern as for the light-field calibration. Stereo calibration is performed between the center sub-aperture $(u,\,v)^T=(5,\,5)^T$ and the infrared camera image.
While we fix the intrinsics of the light-field camera, depth sensor is calibrated only for the intrinsic parameters (no distortion).
After the calibration procedure, we register the depth maps onto the center sub-aperture images. As one can observe in the examples in~\refFig{fig:refocus}, due to the RGB-D sensor noise and the calibration procedure, some pixels around object boundaries do not contain depth measurements (represented in dark blue). Recorded depth maps can be improved further for a better domain adaptation~\cite{liuCVPR13,parkICCV11,parkTIP14,shenCVPR13}. We leave the possible improvements for future work. We convert depth to disparity to generalize the method to different cameras.\\

\noindent{\bf DDFF 12-Scene benchmark.} We collect the dataset in twelve different indoor environments: \textit{glassroom, kitchen, office41, seminar room, social corner, student laboratory, cafeteria, library, locker room, magistrale, office44} and \textit{spencer laboratory}. First six scenes are composed of 100 light-field images and depth pairs and the latter six scenes are composed of 20 pairs. Our scenes have at most 0.5 pixel disparity while the amount of measured disparity gradually decreases towards far distances.~\refFig{fig:dispwhisker} plots the Whisker diagrams for each scene.  Example center sub-aperture images for \textit{office41} and \textit{locker room} scenes and their corresponding disparity maps are shown in~\refFig{fig:refocus}. Since the dataset consists mainly of indoor scenes, flat surfaces (wall, desk), textureless objects (monitor, door, cabinet) and glossy materials (screen, windows) are often present. Our dataset is therefore more challenging and 25 times larger than previous synthetic datasets~\cite{honauer16benchmark,wanner13datasets}. DDFF 12-Scene dataset, consisting of the light-field images, generated focal stacks, registered depth maps and the source code of our method are publicly available on~\url{https://vision.cs.tum.edu/data/datasets/ddff12scene}\,.

\section{Depth From Focus using Convolutional Neural Networks}

This section describes our method for depth reconstruction from a focal stack. We formulate the problem as a minimization of a regression function, which is an end-to-end trained convolutional neural network.

Let $\mathcal{S}$ be a focal stack consisting of $S$ refocused images $I \in \mathbb{R}^{H\times W \times C}$ and the corresponding target disparity map $D \in \mathbb{R}^{H\times W}$. We minimize the least square error between the estimated disparity $f(\mathcal{S})$ and the target $D$:
\begin{equation}
\mathcal{L} = \sum_{p}^{HW} \mathcal{M}(p) \cdot \big \lVert f_\textbf{W}(\mathcal{S}, p) - D(p) \big \rVert_2^{2} + \lambda \lVert \textbf{W} \rVert_2^2\, .
\end{equation}
Loss function $\mathcal{L}$ is summed over all valid pixels $p$ where $D(p) > 0$, indicated by the mask $\mathcal{M}$ and $f:\mathbb{R}^{S\times H\times W \times C} \to\mathbb{R}^{H\times W}$ is a convolutional neural network. Weights, \textbf{W}, are penalized with $\ell_2$-norm. Depth/disparity maps captured by RGB-D sensors often have missing values, indicated with a value of $0$. Therefore, we ignore the missing values during training in order to prevent networks from outputting artifacts.\\

\noindent{\bf Network architecture.} We propose an end-to-end trainable auto-encoder style convolutional neural network. CNNs designed for image classification are mostly encoder type networks which reduce the dimension of the input to a 1D vector~\cite{he16resnet,krizhevsky12alexnet,simonyan15vgg}. This type of networks are very powerful at constructing descriptive hierarchical features later used for image classification. This is why for tasks which require a pixel-wise output, the encoder part is usually taken from these pre-trained networks~\cite{he16resnet,krizhevsky12alexnet,simonyan15vgg} and a mirrored decoder part is created to upsample the output to image size. We follow this same paradigm of hierarchical feature learning for pixel-wise regression tasks~\cite{dosovitsky15flownet,kendall15bayesiansegnet,long15fcn,noh15deconvnet} and design a convolutional auto-encoder network to generate a dense disparity map (see Fig.1 in the supplementary material).

As a baseline for the encoder network, we use the VGG-16 net~\cite{simonyan15vgg}. It consists of 13 convolutional layers, 5 poolings and 3 fully-connected layers. In order to reconstruct the input size, we remove the fully-connected layers and reconstruct the decoder part of the network by mirroring the encoder layers. We invert the $2\times2$ pooling operation with $4\times4$ upconvolution (deconvolution)~\cite{long15fcn} with a stride of $2$ and initialize the weights of the upconvolution layers with bilinear interpolation, depicted as upsample (see supp. Fig. 1).

Similar to the encoder part, we use convolutions after upconvolution layers to further sharpen the activations. To accelerate the convergence, we add batch normalization~\cite{ioffe15batchnorm} after each convolution and learn the scale and shift parameters during training. Batch normalization layers are followed by rectified linear unit (ReLU) activations. Moreover, after the 3rd, 4th and 5th poolings and before the corresponding upconvolutions, we apply dropout with a probability of 0.5 during training similar to~\cite{kendall15bayesiansegnet}. In order to preserve the sharp object boundaries, we concatenate the feature maps of early convolutions \texttt{conv1\_2, conv2\_2, conv3\_3} with the decoder feature maps: outputs of the convolutions are concatenated with the outputs of corresponding upconvolutions (see supp. Fig. 1). 

We refer to this architecture as \textit{DDFFNet}. There are several architectural choices that one can make that can significantly increase or decrease the performance of auto-encoder networks. Some of these changes are the way upsampling is done in the decoder part or the skip connections. For the problem of DFF, we study the performance of the followings variants:
\begin{itemize}
	\item \textit{DDFFNet-Upconv}: In the decoder part, we keep the upconvolutions.
	\item \textit{DDFFNet-Unpool:} Upconvolutions are replaced by $2\times2$ unpooling operation~\cite{dosovitskiy15unpool}.
	\item \textit{DDFFNet-BL:} Upconvolutions are replaced by $2\times2$ bilinear interpolation (upsampling). 
	\item \textit{DDFFNet-CCx:} Here we study the effect of several concatenation connections, designed to obtain sharper edges in the depth maps.    
\end{itemize}

\noindent \textbf{Network input.} VGG-16 net takes the input size of $H\times W\times C$, precisely $224\times 224\times3$. In contrast, we need to input the whole focal stack $\mathcal{S}$ into the network.  Computing features per image $I$ in the focal stack is a general way of incorporating sharpness into DFF approaches~\cite{moeller15vdff} and we make use of this intuition within our end-to-end trained CNN. Since the depth of a pixel is correlated with the sharpness level of that pixel and the convolutions are applied through input channels $C$, we consider the network as a feature extractor and therefore, we reshape our input to $(B\cdot S\times C\times H\times W)$ with a batch size of $B$. Hence, the network generates one feature map per image in the stack with a size of $(B\cdot S\times 1\times H\times W)$. In order to train the network end-to-end, we reshape the output feature maps to $(B\times S\times H\times W)$ and apply $1\times 1$ convolution as a regression layer through the stack, depicted as \textit{Score} in supp. Fig. 1.

\section{Experimental Evaluation}
\label{sec:experiments}
We evaluate our method on the DDFF 12-Scene dataset proposed in~\refSec{sec:sixscene}. We first split the twelve scenes into training and test sets. We use the six scenes,~\ie \textit{cafeteria, library, locker room, magistrale, office44, spencer laboratory} for testing as these scenes have in total 120 focal stacks and are also a good representation of the whole benchmark, as shown in~\refFig{fig:dispwhisker}. The other six scenes are then used for training with a total of 600 focal stacks.\\

\noindent{\bf Evaluation metrics.}
Following~\cite{eigen14depthmap,gargeccv2016,honauer16benchmark,liu16learningdepth} we evaluate the resulting depth maps with eight different error metrics:
\begin{itemize}
	\item MSE : $ \frac{1}{|\mathcal{M}|} \sum_{p \in \mathcal{M}} \big \lVert f(\mathcal{S}_p) - D_{p} \big \rVert_2^{2}$
	\item RMS : $ \sqrt{\frac{1}{|\mathcal{M}|} \sum_{p \in \mathcal{M}} \big \lVert f(\mathcal{S}_p) - D_{p} \big \rVert_2^{2}}$
	\item log RMS :$ \sqrt{\frac{1}{|\mathcal{M}|} \sum_{p \in \mathcal{M}} \big \lVert \log f(\mathcal{S}_p) - \log D_{p} \big \rVert_2^{2}}$
	\item Absolute relative :$\frac{1}{|\mathcal{M}|} \sum_{p \in \mathcal{M}} \frac{ \left | f(\mathcal{S}_p) - D_{p} \right | } { D_{p}}$
	\item Squared relative :$ \frac{1}{|\mathcal{M}|} \sum_{p \in \mathcal{M}} \frac{ \big \lVert f(\mathcal{S}_p) - D_{p} \big \rVert_2^{2} } { D_{p}}$
	\item Accuracy: \% of $D_p$ s.t $\max \big( \frac{f(\mathcal{S}_p)}{D_p}\, , \frac{D_p}{f(\mathcal{S}_p)} \big)= \delta < thr$
	\item BadPix($\tau$): $\frac{\, \left |\{p \in \mathcal{M}\, : \left |f(\mathcal{S}_p) - D_{p} \right | > \tau\} \right |}{|\mathcal{M}|}\cdot 100$
	\item Bumpiness: $\frac{1}{|\mathcal{M}|}\, \sum_{p \in \mathcal{M}}\, \min (0.05, \lVert \text{H}_{\Delta}(p) \rVert_F) \cdot 100$
\end{itemize}  where $\Delta = f(\mathcal{S}_p) - D_{p}$ and \text{H} is the Hessian matrix. The first five measures are standard error measures, therefore lower is better while for the Accuracy measure higher is better. BadPix($\tau$) quantifies the number of wrong pixels with a given threshold $\tau$ while Bumpiness metric focuses on the smoothness of the predicted depth maps~\cite{honauer16benchmark}.\\

\begin{figure}[t!]
	\centering
	\begin{tabular}{cccccccc}
		Image & Disparity & Unpool & BL & UpConv & CC1 & CC2 & CC3	\\ \toprule
		\includegraphics[width=0.115\textwidth,trim={0cm 4cm 5cm 0}, clip]{figures/magistrale_IMG_0012}&
		\includegraphics[width=0.115\textwidth,trim={0cm 4cm 5cm 0}, clip]{figures/magistrale_DISP_0012_gt}&
		\includegraphics[width=0.115\textwidth,trim={0cm 4cm 5cm 0}, clip]{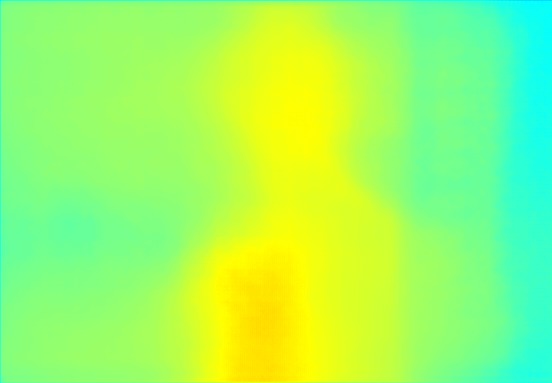}&
		\includegraphics[width=0.115\textwidth,trim={0cm 4cm 5cm 0}, clip]{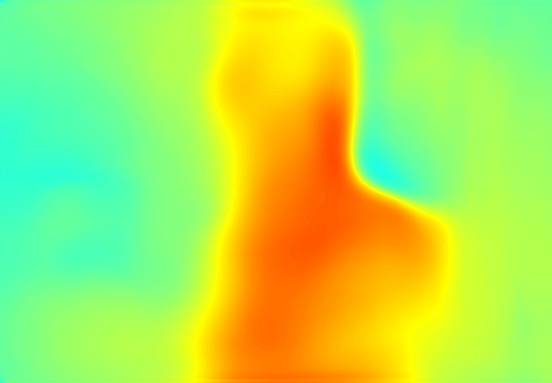}&
		\includegraphics[width=0.115\textwidth,trim={0cm 4cm 5cm 0}, clip]{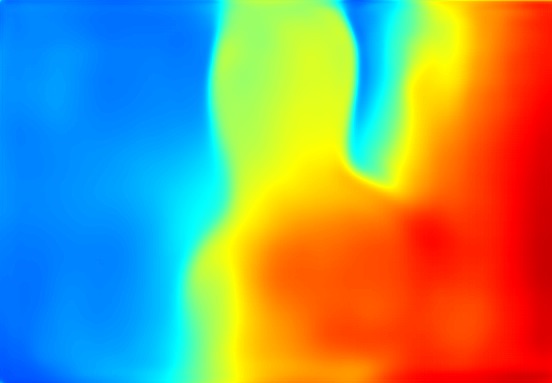}&
		\includegraphics[width=0.115\textwidth,trim={0cm 4cm 5cm 0}, clip]{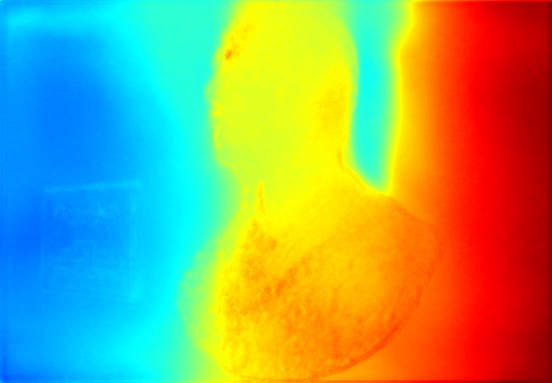}&
		\includegraphics[width=0.115\textwidth,trim={0cm 4cm 5cm 0}, clip]{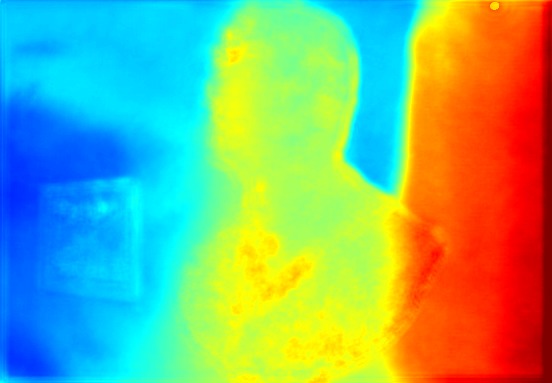}&
		\includegraphics[width=0.115\textwidth,trim={0cm 4cm 5cm 0}, clip]{figures/magistrale_DISP_0012_colored}\\
	\end{tabular}
	\caption{\small \textbf{Qualitative results of the variants of DDFFNet.} While BL oversmooths the edges, CC1 and CC2 introduce artificial edges on the disparity map. Unpooling is not suitable well to recover the fine edges. Best viewed in color.}
	\label{fig:qualResDDFFNet}
\end{figure}

\noindent{\bf Experimental setup.}
For our experiments, we generate the focal stacks for $S=10$ with disparities linearly sampled in $[0.28, 0.02]$ pixel (equivalent to $[0.5, 7]$  meters). We found this to be a good compromise between obtaining pixel sharpness at all depths and memory consumption and runtime, which heavily increases for larger focal stacks without bringing improved depth accuracy (see Figure~\ref{fig:badpix}). 

DDFF 12-Scene consists of $383\times552$ images, thus training on full resolution stacks is inefficient. One solution would be to downsample the images, however, interpolation could change the blur kernels, eventually affecting network performance.
The solution we adopt is to train the network on $10\times 224\times 224\times 3$ stack patches. To do so, we crop the training stacks and corresponding disparity maps with a patch size of 224 and a stride of 56, ensuring that cropped patches cover the whole image. Patches with more than 20\% missing disparity values are removed from the training set. 20\% of the training data is used as validation for model selection. At test time, results are computed on the full resolution $383\times552$ images.

We run all experiments on an NVidia Pascal Titan X GPU. Encoder part of the network is initialized from the pre-trained VGG-16 net, decoder part is initialized with variance scaling~\cite{he15delvingdeep}. We use the SGD optimizer with momentum decay of 0.9. Batch size $B$ is set to 2 and learning rate to 0.001. Every fourth epoch we exponentially reduce the learning rate by a factor of 0.9.

\subsection{Ablation studies} 
We first evaluate our architecture variations such as the three upsampling layers: unpooling, upconvolution and bilinear interpolation. We can see from~\refFig{fig:qualResDDFFNet}, \textit{Unpool} does not preserve the fine object edges while \textit{BL} oversmooths them due to naive linear interpolation. 
These observations are also supported by the quantitative experiments in~\refTab{tab:quanRes}.
Hence, we choose to use \textit{UpConv} for the rest of architectural variations.

Within the tested concatenation schemes, \textit{DDFFNet-CC1} and \textit{DDFFNet-CC2} preserve too many edges as they benefit from larger feature maps. However, this produces incorrect depth and therefore achieving overall worse MSE compared to that of \textit{DDFFNet-CC3}, see~\refTab{tab:quanRes}. On the other hand, \textit{DDFFNet-CC3} preserves only the most important edges corresponding to object boundaries. Going deeper in the concat connections would not provide sufficiently fine structures, hence, we do not test connections after \textit{CC3}.

We further plot the BadPix measure when changing the threshold $\tau$ in ~\refFig{fig:badpix}. In this plot, we compare our best architecture \textit{DDFFNet-CC3} with focal stacks of varying sizes, $S \in \{5, 8, 10, 15\}$. Having light-field images allows us to digitally generate focal stacks of varying sizes, enabling us to find an optimal size. Even though increasing the stack size $S$ should theoretically decrease the depth error, $S$-15 quickly overfits due to the fact that it was trained with a batch size of 1 to fit into the memory of a single GPU. We find that a focal stack of 10 images is the best memory-performance compromise, which is why all further experiments are done with $S=10$.

\begin{table}[t]
	\centering
	\setlength{\tabcolsep}{3pt}
	\scriptsize
	\begin{tabular}{l l c c c c c c c c c}
		\multicolumn{2}{c}{ } & & & & & &\multicolumn{3}{c}{Accuracy ($\delta=1.25$)} & \\
		\cmidrule(lr){8-10}
		\multicolumn{2}{c}{Method} & MSE $\downarrow$ & RMS $\downarrow$ & $\log$ RMS $\downarrow$& Abs. rel. $\downarrow$& Sqr. rel.$\downarrow$ & $\delta$ $\uparrow$ & $\delta^2$ $\uparrow$ & $\delta^3$ $\uparrow$& Bump. $\downarrow$\\
		\toprule
		& Unpool &2.9$\,e^{-3}$& 0.050&0.50&0.64&0.05&39.95&63.46&78.26&0.62
		\\
		& BL & 2.1$\,e^{-3}$&0.041&0.43&0.46&0.03&51.29&74.81&85.28&0.54
		\\
		\parbox[t]{2mm}{\multirow{4}{*}{\rotatebox[origin=c]{90}{DDFFNet}}}
		& UpConv &1.4$\,e^{-3}$&0.034&0.33&0.30&0.02&52.41&83.09&93.78& 0.54
		\\
		& CC1  & 1.4$\,e^{-3}$& 0.033& 0.33&0.37& 0.02& 60.38& 82.11& 90.63& 0.75
		\\
		& CC2 & 1.8$\,e^{-3}$& 0.039&0.39& 0.39& 0.02& 44.80& 76.27& 89.15 &  0.75
		\\ 
		& CC3 & 9.7$\,e^{-4}$&0.029&0.32&0.29&0.01& 61.95& 85.14& 92.99& 0.59
		\\
		\arrayrulecolor{lightgray}\midrule
		& PSPNet & 9.4$\,e^{-4}$& 0.030& 0.29& 0.27& 0.01& 62.66& 85.90& 94.42& 0.55
		\\
		& Lytro &2.1$\,e^{-3}$& 0.040&0.31&0.26&0.01&55.65&82.00&93.09& 1.02
		\\
		& PSP-LF & 2.7$\,e^{-3}$& 0.046 &	0.45 & 0.46 &	0.03 &	39.70 & 65.56 & 82.46 & 0.54
		\\
		& DFLF & 4.8$\,e^{-3}$& 0.063& 0.59&0.72& 0.07&28.64& 53.55& 71.61&  0.65\\	
		& VDFF & 7.3$\,e^{-3}$& 0.080& 1.39& 0.62& 0.05& 8.42& 19.95& 32.68& 0.79\\
		\arrayrulecolor{lightgray}\midrule
	\end{tabular}
	\caption{\label{tab:quanRes}\small \textbf{Quantitative results.} \textit{DDFFNet-CC3} is the best depth from focus method and provides also better results compared to Lytro,~\ie depth from light-field. Metrics are computed on the predicted and the groundtruth disparity maps.}
\end{table}

\subsection{Comparison to state-of-the-art} 

We have implemented three baselines to compare our method with:\\

\noindent{\bf Variational DFF.} We compare our results with the state-of-the-art variational method, VDFF~\cite{moeller15vdff}, using their GPU code\footnote{\url{https://github.com/adrelino/variational-depth-from-focus}} and the same focal stacks as in our method. We run a grid search on several VDFF parameters and the results reported are for the best set of parameters. VDFF outputs a real valued index map. Each pixel is assigned to one of the stack images, where the pixel is in focus. Therefore, we directly interpolate these indices to their corresponding disparity values and compute our metrics on the mapped disparity output.\\

\noindent{\bf PSPNet for DFF.} \emph{Pyramid Scene Parsing Network}~\cite{zhao17pspnet} is based on a deeper encoder (ResNet) and also capable of capturing global context information by aggregating different-region-based context through the pyramid pooling module. It is originally designed for semantic segmentation, however, we modified the network for depth from focus problem (input and output) and trained it end-to-end on our dataset. We also compare to PSPNet in order to observe the effects of significant architectural changes in terms of a deeper encoder (ResNet) with a recent decoder module for the problem of depth from focus.\\

\noindent{\bf Lytro depth.} For completion, we also compare with the depth computed from the light-field directly by the Lytro. Although this method technically does not compute DFF, we still think it is a valuable baseline to show the accuracy that depth from light-field methods can achieve. Lytro toolbox predicts depth in lambda units, thus the output is not directly comparable to our results. For this reason, we formulate the rescaling from Lytro depth to our groundtruth as an optimization problem that finds a unique scaling factor $k^\ast$.
To do so, we minimize the least squares error between the resulting depth $\tilde{Z}_p$ and the groundtruth depth $Z_p$ to find the best scaling factor $k^\ast$:
\begin{equation}
k^\ast = \arg\, \min_k \sum_p\, \lVert k \cdot \tilde{Z}_p - Z_p \rVert_2^2\, ,
\end{equation} 
where $k \in \mathbb{R}$. Note that this is the best possible mapping in terms of MSE to our groundtruth depth maps provided that the focal stack has uniform focal change, therefore, we are not penalizing during the conversion process. Evaluation metrics are then computed on $k^\ast \cdot \tilde{D}_p$ and $D_p$.\\

\noindent{\bf Depth from light-field.} Even though we focus on the task of DFF, we want to provide a comparison to depth from light-field. For this purpose, we follow~\cite{heberCVPR16,heberICCV17} and train our network (\textit{DDFFNet-CC3}) as well as PSPNet with 11 sub-apertures from the light-field camera as input (see supp. Fig. 3). We denote these models as DFLF and PSP-LF, respectively. 
To the best of our knowledge, there is no code for~\cite{heberCVPR16,heberICCV17} to test their full pipeline on our dataset. 

As we can see from~\refTab{tab:quanRes}, \textit{DDFFNet-CC3} outperforms the other depth from focus method,~\ie VDFF~\cite{moeller15vdff}, in all evaluation metrics, reducing depth error by more than 75\%.
The major reason for this is that VDFF proposes an optimization scheme that relies on precomputed hand-crafted features, which can handle the synthetic or clean high resolution images but fail in the realistic, challenging scenes of our dataset.

PSPNet performs on-par to \textit{DDFFNet-CC3}, nevertheless, as shown in~\refFig{fig:qualRes}, pyramid pooling leads to oversmooth depth maps due to its upsampling strategy. Although this network is very efficient for semantic segmentation, we found that our decoder choice with skip connection \text{CC3} yields more accurate, non-smooth depth maps.

\begin{figure}[t!]
	\begin{minipage}{.45\textwidth}
		\centering
		\setlength{\tabcolsep}{1pt}
		\begin{tabular}{l l c  c}
			\multicolumn{2}{c}{Method}  & Runtime (s.) & Depth (m.) \\
			\toprule
			& Unpool & 	0.55 & 1.40\\
			& BL  & \textbf{0.43} & 1.10\\
			\parbox[t]{3mm}{\multirow{4}{*}{\rotatebox[origin=c]{90}{DDFFNet}}}
			& UpConv &	0.50 & \textbf{0.58}\\ 
			& CC1 & 0.60 & 0.79\\
			& CC2 & 0.60  & 0.86\\
			& CC3 & 0.58 & 0.86\\
			\arrayrulecolor{lightgray}\midrule
			& DFLF & 0.59& 1.50\\
			& VDFF & 2.83 & 8.90\\ \midrule
			& Lytro & 25.26 (CPU)  & 0.99 
		\end{tabular}
		\captionof{table}{\small \label{tab:runtime}\textbf{Runtime and Depth error.} DDFFNet is faster and more accurate than other state-of-the-art methods. For completeness, we also report the runtime of Lytro toolbox on CPU. VDFF performs worse as it requires many iterations of optimization during test.}
	\end{minipage}
	\quad
	\begin{minipage}{0.45\textwidth}
		\centering
		\includegraphics[width=\textwidth]{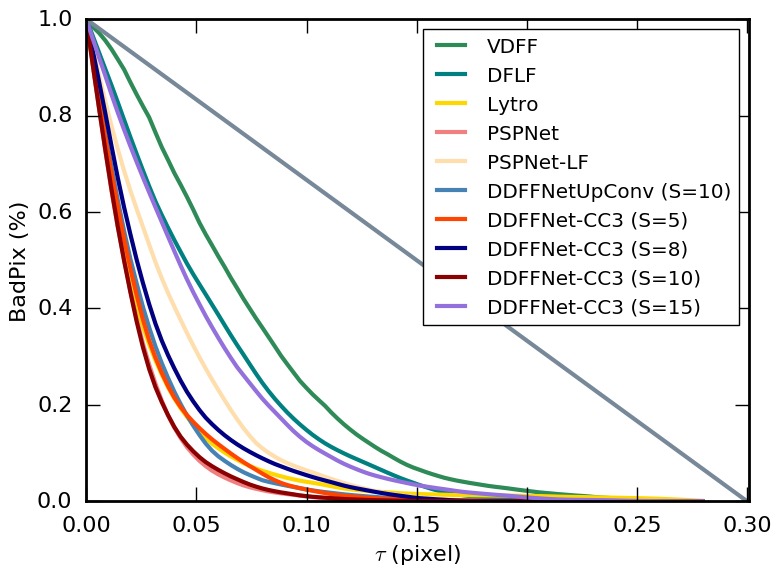}
		\captionof{figure}{\small \label{fig:badpix}\textbf{Badpix(\%)} for DDFFNet-CC3 for $S \in \{5, 8, 10, 15\}$, DDFFNet-UpConv for $S$=$10$, for Lytro, DFLF and VDFF. While $\tau$ increases BadPix error decreases. DDFFNet with stack size of 10 is better than VDFF and Lytro by a large margin.}
	\end{minipage}
\end{figure}

Lytro, on the other hand, computes very inaccurate depth maps on flat surfaces (\refFig{fig:qualRes}) while our \textit{DDFFNet} estimates a smoother and also more accurate disparity maps. Note that Lytro depth is what is provided as groundtruth depth maps in the Stanford Light-field dataset, but this depth computation is relying on an algorithm that is not always accurate as we have shown. This is another advantage of the proposed dataset which provides groundtruth maps from an external RGB-D sensor.

As we can see in ~\refTab{tab:quanRes}, our approach \textit{DDFFNet-CC3} still performs better than depth from light-field (DFLF) with a similar number of images. Note that it would not be possible to fit all sub-aperture images provided by the light-field camera into GPU memory. In contrast to DFLF, our method is usable with any camera. 

Moreover, we present the MSE and RMS error computed on the predicted depth maps in~\refTab{tab:runtime}. \textit{DDFFNet-CC3} achieves a much lower error when compared to VDFF or DFLF. \textit{DDFFNet-UpConv} has a better depth error than \textit{DDFFNet-CC3}, but, its badpix error is significantly larger than \textit{DDFFNet-CC3}, demonstrated in~\refFig{fig:badpix}.

Overall, experiments show that our method is more accurate by a large margin when compared to the classical variatonal DFF method~\cite{moeller15vdff} while also being orders of magnitude faster on a GPU.
It is also more accurate than the Lytro predicted depth or a network trained for depth from light-field.
Several network architectures were explored, and finally \textit{CC3} was deemed the best with overall lowest disparity error while keeping object boundaries in the disparity map.\\

\noindent \textbf{Is DDFFNet generalizable to other cameras?}
To show the generality of our method, we capture focal stacks with an Android smartphone where the focus changes linearly from $0.1$ to $4m$ and groundtruth depth maps with the depth camera of the smartphone. We present the results on this \textit{mobile} depth from focus (mDFF) dataset in the supplementary material and share the dataset publicly on~\url{https://vision.cs.tum.edu/data/datasets/mdff}\,.

\section{Conclusions}

Depth from focus (DFF) is a highly ill-posed inverse problem because the optimal focal distance is inferred from sharpness measures which fail in untextured areas.  Existing variational solutions revert to spatial regularization to fill in the missing depth, which are not generalized to more complex geometric environments.  
In this work, we proposed `Deep Depth From Focus' (DDFF) as the first deep learning solution to this classical inverse problem.  To this end, we introduced a novel 25 times larger dataset with focal stacks from a light-field camera and groundtruth depth maps from an RGB-D camera.  We devised suitable network architectures and demonstrated that DDFFNet outperforms existing approaches, reducing the depth error by more than 75\% and predicting a disparity map in only 0.58 seconds. 

\newpage
\bibliographystyle{splncs04}
\bibliography{ddff}



\title{Deep Depth From Focus\\ {\small Supplementary Material}} 
\titlerunning{Deep Depth From Focus {\small Supplementary Material}} 


\author{Caner Hazirbas \and Sebastian Georg Soyer \and Maximilian Christian Staab\and\\ Laura Leal-Taix\'e \and Daniel Cremers}
%
\index{Soyer, Sebastian Georg}
\index{Staab, Maximilian Christian}

\authorrunning{C. Hazirbas et al.} 


\institute{Technical University of Munich, Germany\\
	\email{\{hazirbas, soyers, staab, leal.taixe, cremers\}@cs.tum.edu}}
\maketitle
\setcounter{page}{1}
\setcounter{section}{0}
\setcounter{figure}{0}
\setcounter{table}{0}
\setcounter{equation}{0}

\begin{abstract}
	In this supplementary material we include additional information for the reader. 
	We first detail the formulation of the light-field calibration and provide the values for all parameters.
	We then present the characteristics of the new DDFF 12-Scene dataset, such as disparity histograms per sequence and disparity sampling. Moreover, we show more qualitative results including the failure cases and also visualizations of the activation heat maps for our best performing model,~\ie \textit{DDFFNet-CC3}. Finally, we provide the results on the 4D light-field dataset and also on our \textit{mobile} DFF (mDFF) dataset.
\end{abstract}
\section{Light-field Camera Calibration}
We make use of the light-field camera calibration toolbox by Bok~\etal~\cite{bokpami2017}, which generates the sub-apertures based on a \textit{radius} $r_m$ of a microlens image, which is set to 7 pixels for the Lytro ILLUM camera. Although the toolbox generates 13$\times$13 sub-apertures, we follow the authors' recommendation~\cite{bokpami2017} to only use the sub-apertures within the displacement $i^2 + j^2 < (\textit{radius}-1)^2$. 
This results in 9$\times$9 sub-apertures. Estimated intrinsic parameters of the Lytro ILLUM are given in~\refTab{tab:intrinsics}. Intrinsics of the microlenses are computed as
\begin{equation}
\text{Int} = \begin{bmatrix}
F_x / (2r_m) & 0 & C_x / (2r_m)\\
0 & F_x / (2r_m) & C_y / (2r_m)\\
0 & 0 & 1
\end{bmatrix}.
\end{equation}

\begin{figure}[!t]
	\centering
	\includegraphics[width=\textwidth]{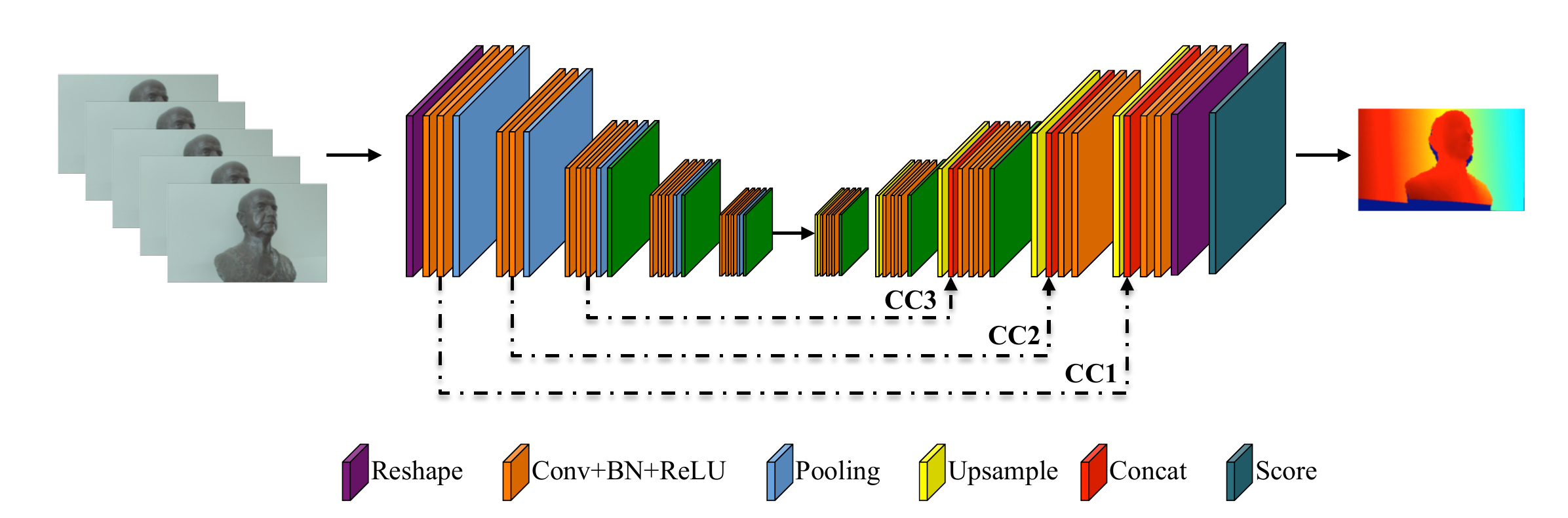}
	\caption{\small \textbf{DDFFNet.} Proposed auto-encoder-style architecture that takes in a focal stack and produces a disparity map. We present several architectural modifications, namely CC connections, Upsample,~\ie Unpool, BL and UpConv.}
	\label{fig:ddffnet}
\end{figure}

\begin{table}
	\centering
	\setlength{\tabcolsep}{10pt}
	\begin{tabular}{lc|lc}
		\multicolumn{4}{c}{Parameters of the Lytro ILLUM}\\
		\toprule
		$r_m$ & 7 &  $F_x$ & 7299.7 \\
		$K_1$ & -2.768 & $F_y$ & 7317.0\\
		$K_2$ & 1982.0 & $C_x$ & 3991.6 \\
		$k_1$ & 0.388 & $C_y	$ & 2629.6\\
		$k_2$ & -0.0361	& $K_1 / F'$ & 27$\mathrm{e}{-5}$
	\end{tabular}
	\caption{\label{tab:intrinsics}\small\textbf{Estimated intrinsic} parameters of the Lytro ILLUM. $F_\ast$ and $C_
		\ast$ are respectively the focal length and optical center of the main lens in pixels. Baseline ($K_1 / F'$) is the distance between two adjacent sub-apertures in meter/pixel, where $F'=\max(F_x, F_y)$. Refer to~\cite{bokpami2017} for details.}	
\end{table}

\section{DDFF 12-Scene Dataset}
Our new dataset is composed of 12 scenes. We use as training set the first six scenes with 100 light-field samples each, for a total of 600 light-field training images. The other six scenes have 20 light-field samples each and are used for testing. All 720 light-field images have registered groundtruth depth/disparity maps obtained from an RGB-D sensor.

In ~\refFig{fig:disphist}, the disparity distribution of the twelve scenes is shown. In ~\refFig{fig:disphist_norm}, we plot the normalized disparity histogram of the training and test sets. We generate the focal stacks for 10 sampled disparities in the interval of $[0.28, 0.02]$ pixels (equivalent to $[0.5, 7]$ meters), indicated with blue dashed lines in~\refFig{fig:disphist_norm}. We also plot the depth to disparity conversion for the given baseline and focal length of the microlenses in~\refFig{fig:refocusdisp}. Refocused disparity values and their corresponding depths are denoted with a green box. Note that disparity is inversely proportional to depth and therefore, linear sampling in disparity corresponds to non-linear sampling in depth. 
We choose to sample disparities to have a linear focus change between stack images.

\section{DDFFNet results}
We present the network architecture in~\refFig{fig:ddffnet} and further qualitative results in~\refFig{fig:qualRes}. Note the poor performance of classic methods like VDFF, and even the wrong disparity maps produced by Lytro in the first row. In ~\refFig{fig:qualResFail}, we present two failure cases where the network output is not sharp or not consistent. 

\begin{figure}
	\def\colsep{1pt}
	\def\rowstretch{0.5}
	\centering
	\begin{tabular}{cccccc}
		\includegraphics[width=0.16\textwidth,trim={0cm 4cm 5cm 0}, clip]{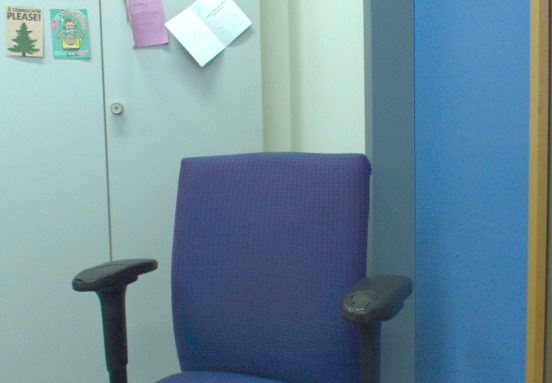}&
		\includegraphics[width=0.16\textwidth,trim={0cm 4cm 5cm 0}, clip]{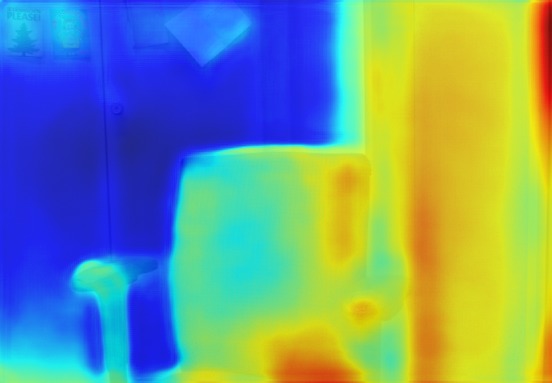}&
		\includegraphics[width=0.16\textwidth,trim={0cm 4cm 5cm 0}, clip]{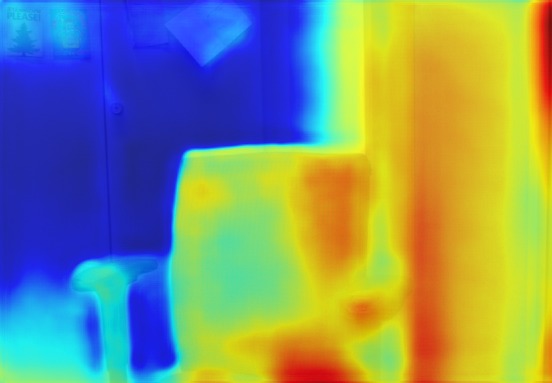}&
		\includegraphics[width=0.16\textwidth,trim={0cm 4cm 5cm 0}, clip]{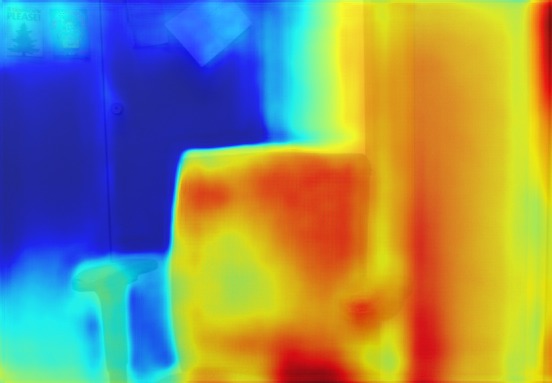}&
		\includegraphics[width=0.16\textwidth,trim={0cm 4cm 5cm 0}, clip]{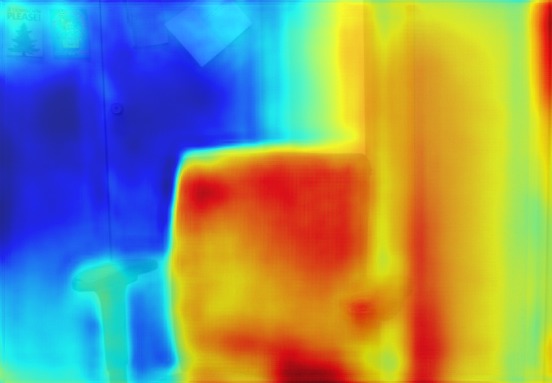}&
		\includegraphics[width=0.16\textwidth,trim={0cm 4cm 5cm 0}, clip]{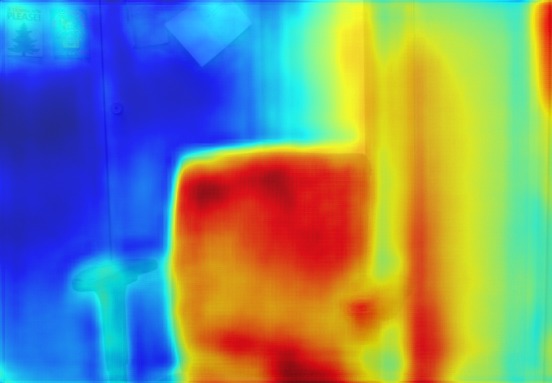}\\
		\includegraphics[width=0.16\textwidth,trim={0cm 4cm 5cm 0}, clip]{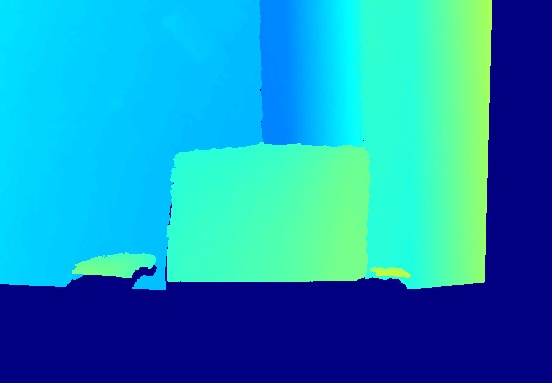}&
		\includegraphics[width=0.16\textwidth,trim={0cm 4cm 5cm 0}, clip]{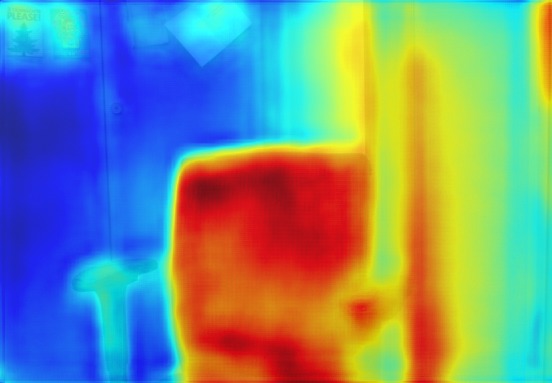}&
		\includegraphics[width=0.16\textwidth,trim={0cm 4cm 5cm 0}, clip]{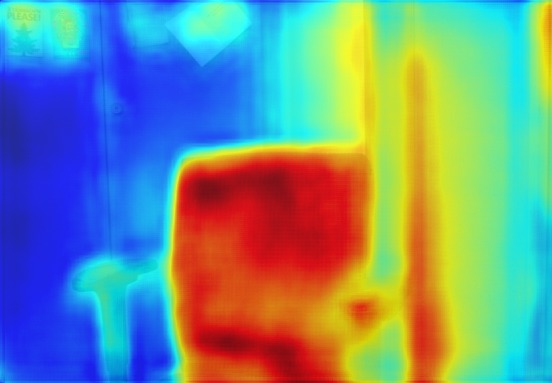}&
		\includegraphics[width=0.16\textwidth,trim={0cm 4cm 5cm 0}, clip]{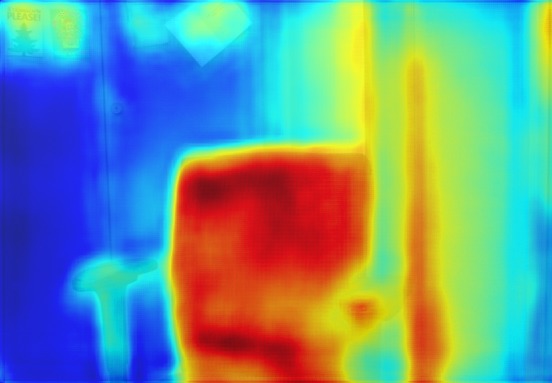}&
		\includegraphics[width=0.16\textwidth,trim={0cm 4cm 5cm 0}, clip]{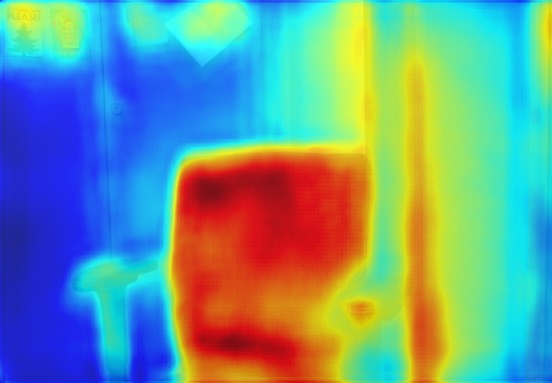}&
		\includegraphics[width=0.16\textwidth,trim={0cm 4cm 5cm 0}, clip]{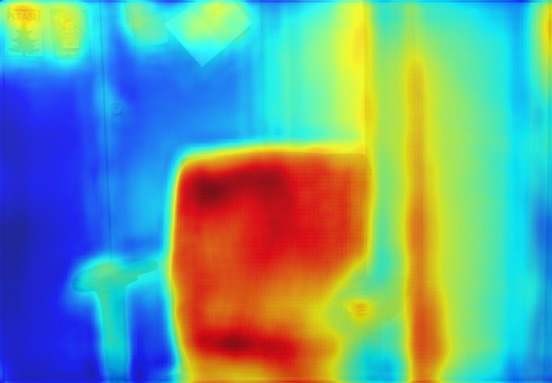}
	\end{tabular}
	\caption{\small \textbf{Activation heat maps} for the refocused images in a focal stack. First column: {\it top:} center sub-aperture image, {\it bottom:} groundtruth disparity map. The rest of the columns show from left to right, top to bottom, how the activations on the focal stack images evolve. Note how the activations slowly shift to the closest object (the chair) as we advance in the focal stack.}
	\label{fig:activmaps2}
\end{figure}

\paragraph{\bf What is the network learning?}
Following the convention in depth from focus, one can see the DDFFNet as a sharpness measure. Our network takes an input of $B \times S$ and reshapes it before the first convolution. This design allows the network to learn in which image of the stack the pixel is sharpest. From this sharpness level, our regression layer, denoted as \textit{Score}, regresses the depth from sharpness (focus). Proof of the concept is visually provided in~\refFig{fig:activmaps2}, where we can see how the activations move towards the closest object (the chair) as we advance in the focal stack, hinting at the fact that the network is really using the focus information to estimate the depth.

\begin{figure}
	\centering
	\includegraphics[scale=1]{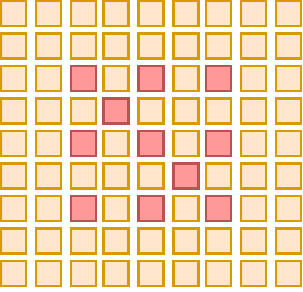}
	\caption{\label{fig:lf_sampling} \textbf{Sampled subpartures for DFLF} are demonstrated with red color.}
\end{figure}

\begin{figure}[t!]
	\def\colsep{1pt}
	\def\rowstretch{0.5}
	\centering
	\begin{tabular}{ccc}
		\multicolumn{3}{c}{\textbf{Training set}} \\ \midrule
		\includegraphics[width=0.32\textwidth]{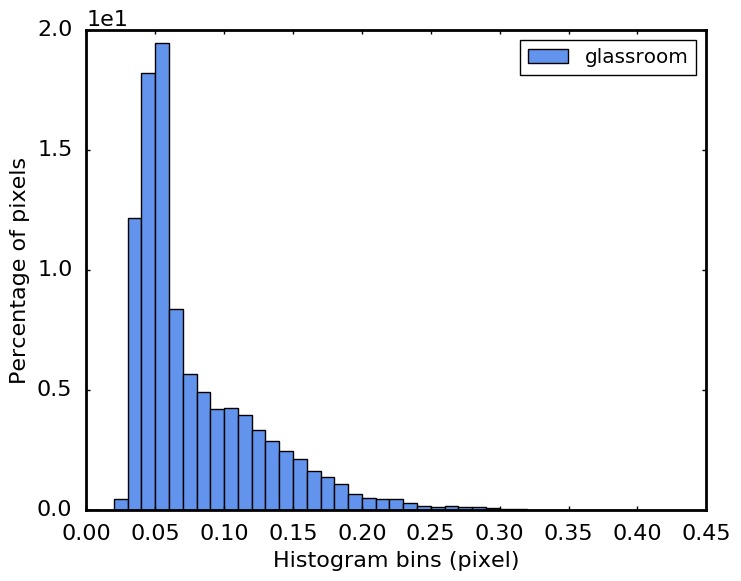} &
		\includegraphics[width=0.32\textwidth]{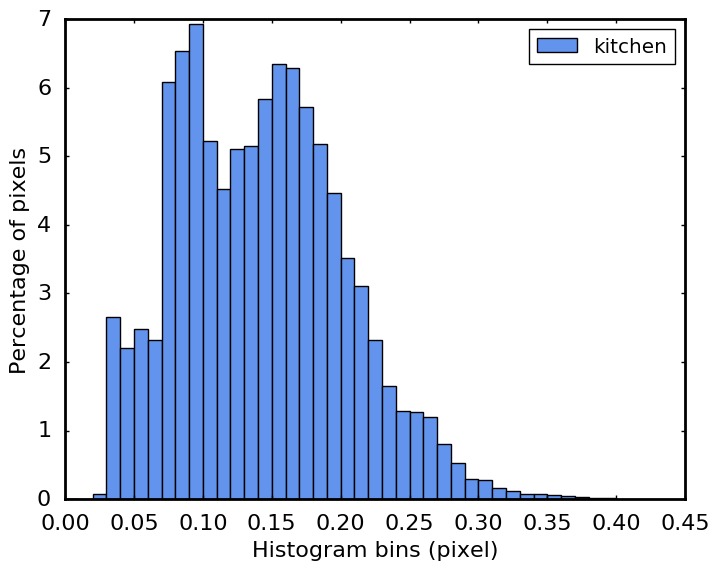} &
		\includegraphics[width=0.32\textwidth]{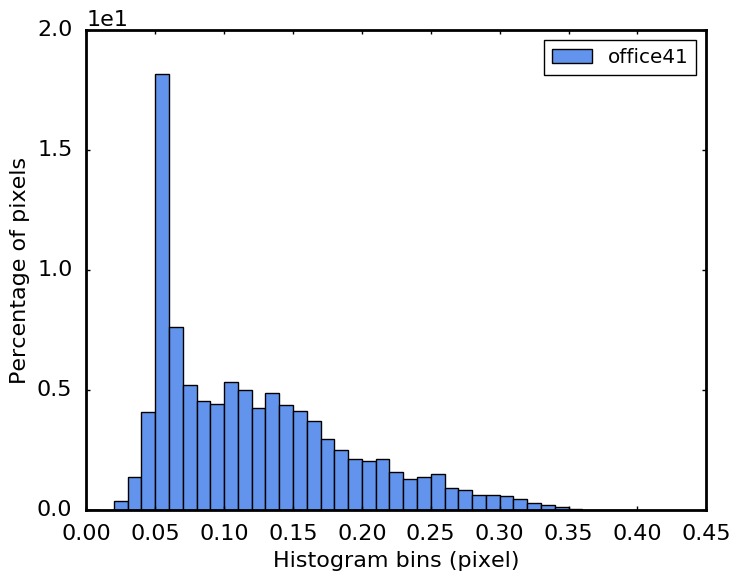} \\
		\includegraphics[width=0.32\textwidth]{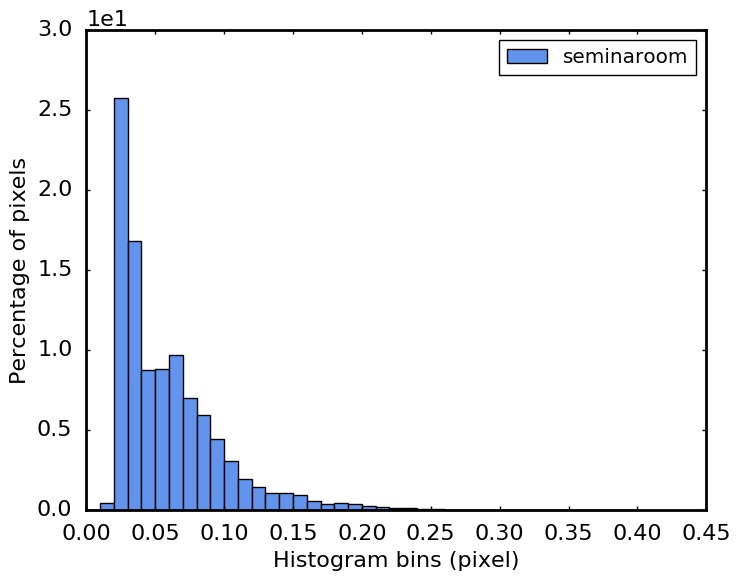} &
		\includegraphics[width=0.32\textwidth]{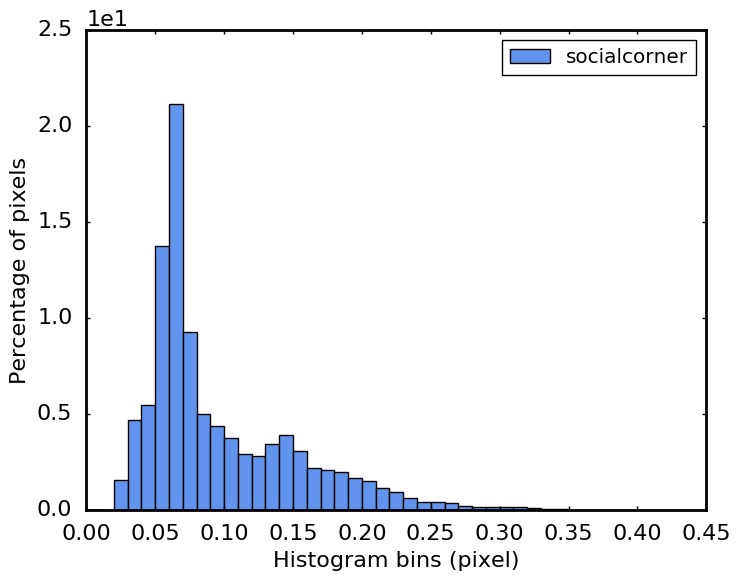} &
		\includegraphics[width=0.32\textwidth]{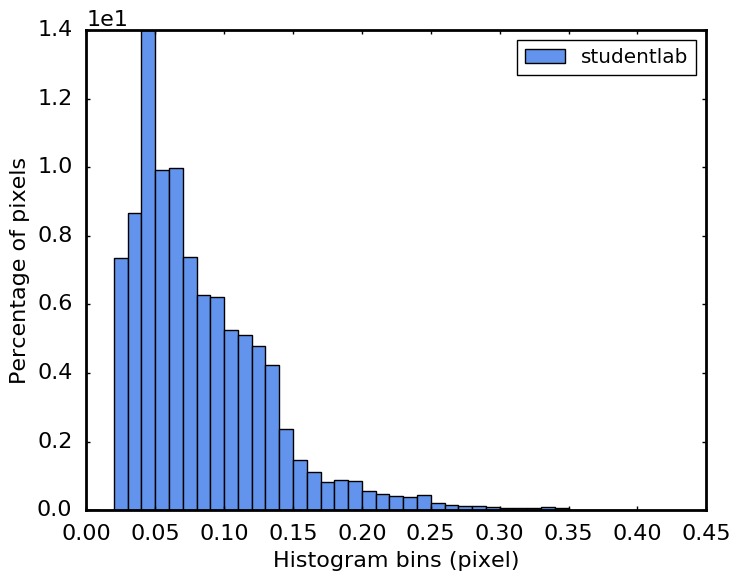}\\ \\
		\multicolumn{3}{c}{\textbf{Test set}} \\ \midrule
		\includegraphics[width=0.32\textwidth]{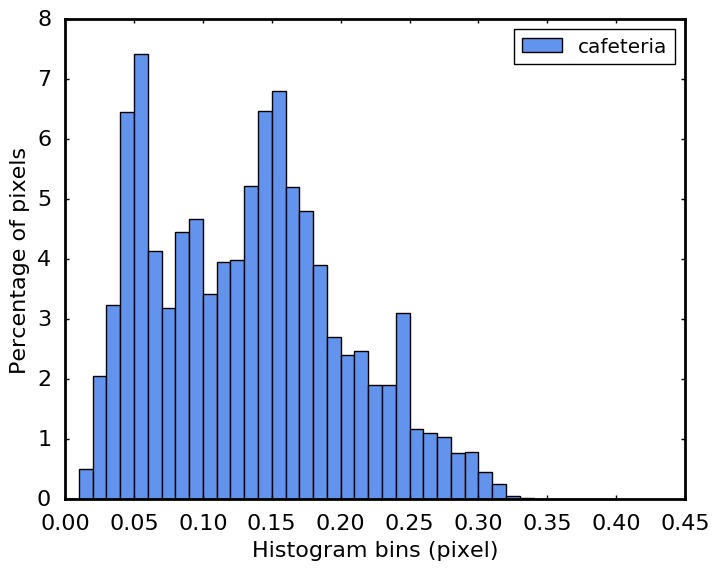} &
		\includegraphics[width=0.32\textwidth]{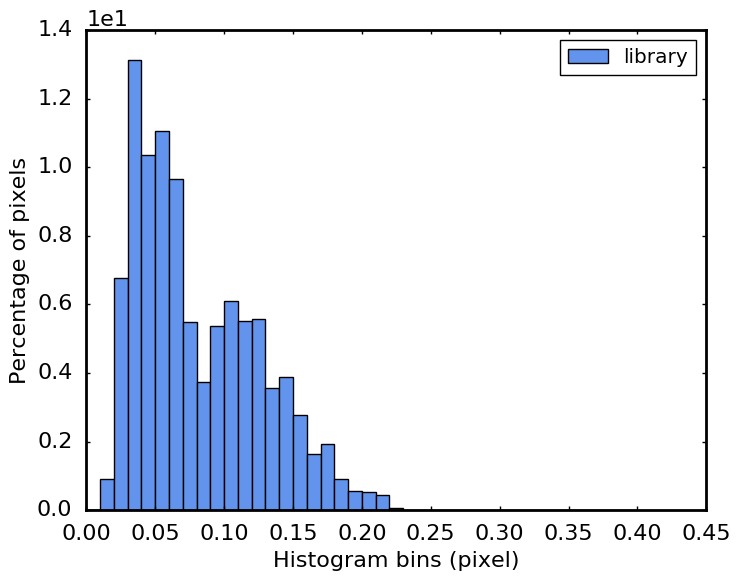} &
		\includegraphics[width=0.32\textwidth]{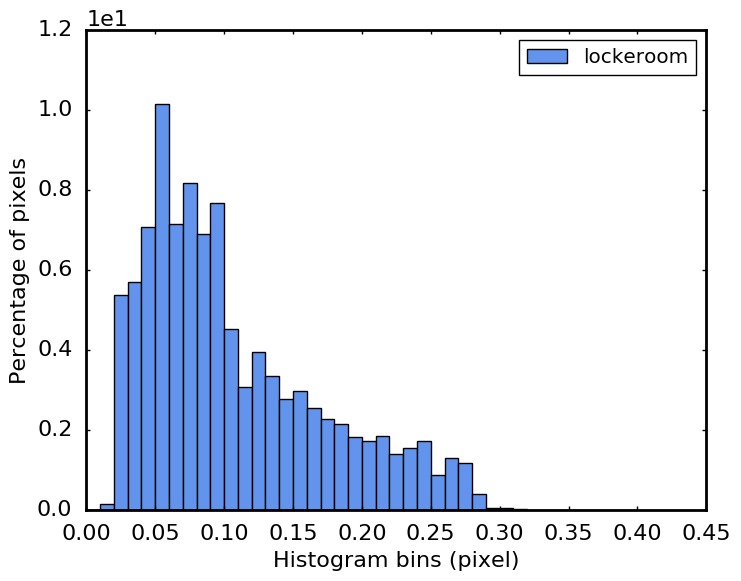} \\
		\includegraphics[width=0.32\textwidth]{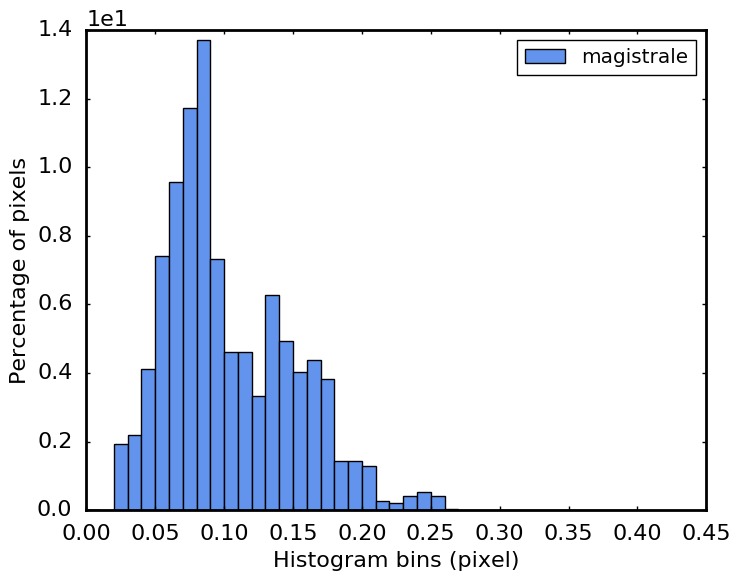} &
		\includegraphics[width=0.32\textwidth]{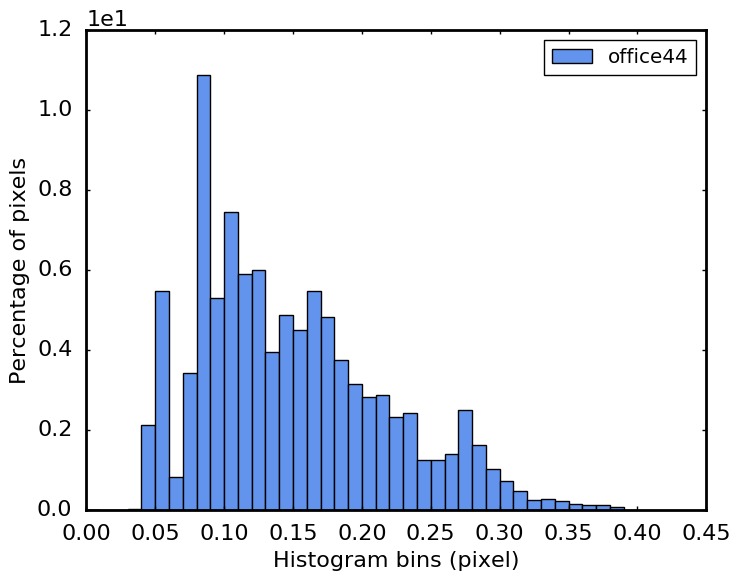} &
		\includegraphics[width=0.32\textwidth]{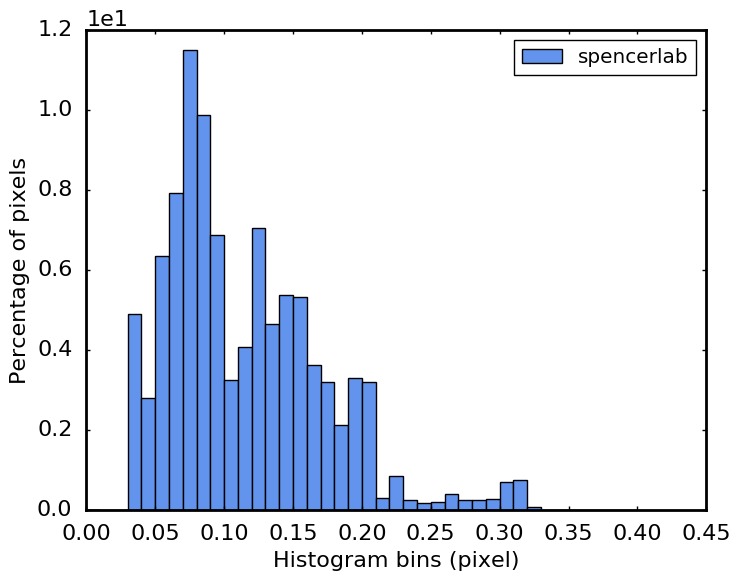}			
	\end{tabular}
	\caption{\small \textbf{Disparity distribution} of the DDFF 12-Scene dataset. Each of the first six scenes is composed of 100 light-field samples and used for training. Each of the latter six scenes is composed of 20 light-field samples and used for testing.}
	\label{fig:disphist}
\end{figure}

\begin{figure}
	\centering
	\begin{subfigure}[t]{0.46\textwidth}
		\includegraphics[width=\textwidth]{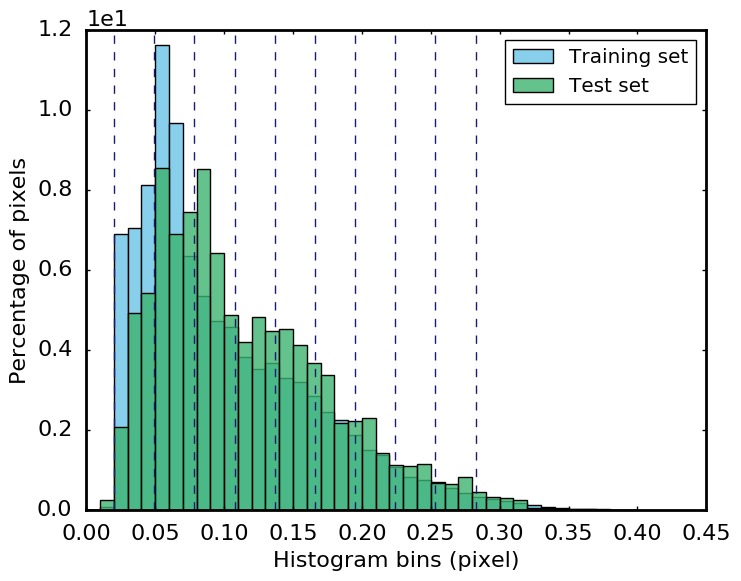}
		\caption{\label{fig:disphist_norm}}
	\end{subfigure}
	\qquad
	\begin{subfigure}[t]{0.44\textwidth}
		\includegraphics[width=\textwidth]{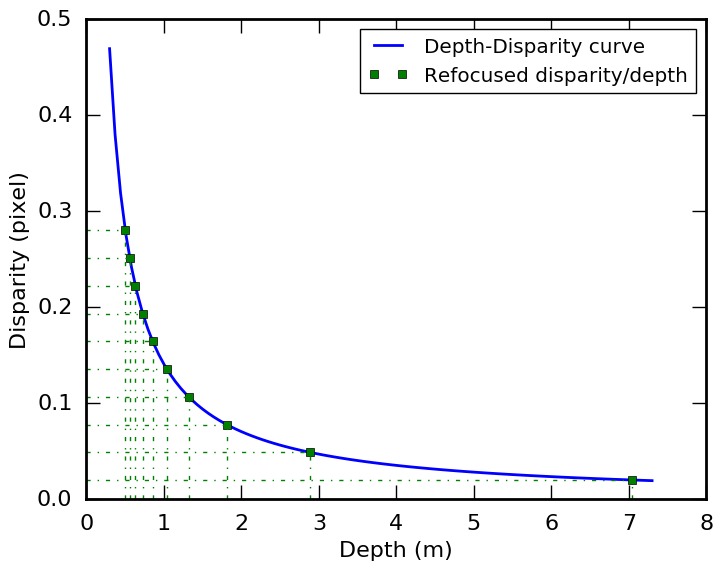}
		\caption{\label{fig:refocusdisp}}
	\end{subfigure}
	\caption{\small \textbf{(a) Disparity distribution} of the training and test sets. Dashed blue lines represent the sampled disparity values used to generate the focal stacks. \textbf{(b) Depth to disparity} conversion for DDFF 12-Scene dataset. Sampled disparities used for refocusing and their corresponding depths are denoted with green boxes.}
	\label{fig:dispdepth}
\end{figure}

\noindent \textbf{LF sampling.} To have a fair comparison we ran the \textit{depth from light-field} experiments with 11 sub-apertures sampled from $9\times9$ grid shown in~\refFig{fig:lf_sampling}. DDFFNet is originally designed to solve the \textit{depth from focus} problem. Nonetheless, one can easily use the same network architecture for DFLF. In \textit{depth from light-field} depth can be recovered by measuring the displacement of pixels from one subparture to another and therefore, one can derive a better network architecture for the \textit{depth from light-field} problem. On the other hand, increasing the number of sampled sub-apertures would also yield better performance.

\section{Results on the 4D light-field dataset}
We also present the results of \textit{DDFFNet-CC3} and PSPNet on the 4D light-field dataset~\cite{honauer16benchmark}. Note that this dataset is not designed for depth from focus but rather depth from light-field. Therefore, we generate a focal stack of refocused images per scene in the disparity interval of~$[-2.5, 2.5]$ from the light-field images. The number of images in this benchmark is limited and training a network from scratch with the provided data does not work. We propose to do transfer learning from our DDFF 12-Scene dataset by fine-tuning \textit{DDFFNet-CC3} and PSPNet.  As training set for fine-tuning we use the 16 light-field images marked as ``additional'' and 4 marked as ``stratified''. As test set we use the 4 scenes marked as ``train'' set. 
\begin{table}[h]
	\centering
	\setlength{\tabcolsep}{4pt}
	\begin{tabular}{l c c c c}
		Method & MSE $\downarrow$ & RMS $\downarrow$ & Bump. $\downarrow$\\
		\toprule
		VDFF & 1.30& 1.15& 1.58\\
		PSPNet & 0.37 & 0.53 & 1.21\\
		DDFFNet-CC3& 0.19& 0.42& 1.92
	\end{tabular}
	\caption{\small \label{tab:4dlfquanRes}\textbf{Quantitative results of the proposed method.} Metrics are computed on the predicted and groundtruth disparity maps.}
\end{table}
In~\refFig{fig:4dlfdataset}, we show the  qualitative results and in Table~\ref{tab:4dlfquanRes}, we present the quantitative results.~\textit{DDFFNet-CC3} outperforms other methods in terms of MSE and RMS error, showing that by using only 20 images from a completely different setting, we can fine-tune our network to achieve accurate results.

\section{Mobile  depth from focus dataset (mDFF)}
We also collected a smaller dataset using a Lenovo Phab 2 Pro smartphone. This phone has a time-of-flight depth sensor that allows us to capture focal stacks with groundtruth \textit{depth} maps.  We collect 202 samples indoor and outdoor. Each sample has 10 refocused images where the focus changes from 0.1 to 4$m$.

To show the performance of our approach, we train the \textit{DDFFNet-CC3} and PSPNet on 181 samples and test on the rest 21 samples. Example images and results are shown in~\refFig{fig:android}. Quantitative comparison is presented in~\refTab{tab:android}.

\begin{table}[h]
	\centering
	\setlength{\tabcolsep}{4pt}
	\begin{tabular}{l c c c c}
		Method & MSE $\downarrow$ & RMS $\downarrow$ & Bump. $\downarrow$\\
		\toprule
		VDFF & 1.00& 0.96& 1.78\\
		PSPNet & 0.46& 0.60& 1.72\\
		DDFFNet-CC3& 0.54 & 0.67 & 2.09 	
	\end{tabular}
	\caption{\small \label{tab:android}\textbf{Quantitative results} on the \textit{mDFF} dataset. Metrics are computed on the predicted and groundtruth \textit{depth} maps.}
\end{table}

\begin{figure}
	\def\colsep{1pt}
	\def\rowstretch{0.5}
	\centering
	\begin{tabular}{c c c c c c c c}
		Image & Disparity & VDFF & DDLF & PSP-LF & Lytro & PSPNet & Proposed
		\\
		\includegraphics[width=0.115\textwidth,trim={0cm 4cm 5cm 0}, clip]{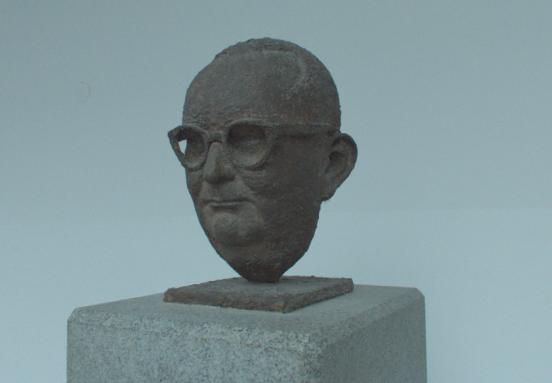}&
		\includegraphics[width=0.115\textwidth,trim={0cm 4cm 5cm 0}, clip]{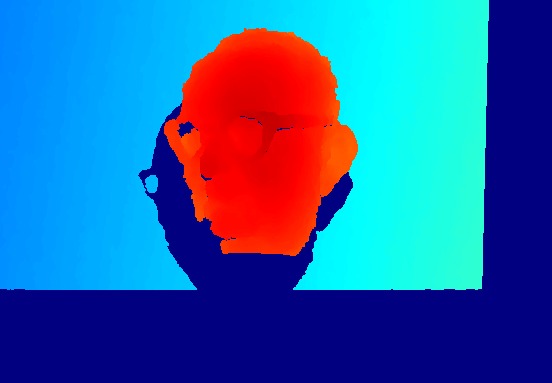}&
		\includegraphics[width=0.115\textwidth,trim={0cm 4cm 5cm 0}, clip]{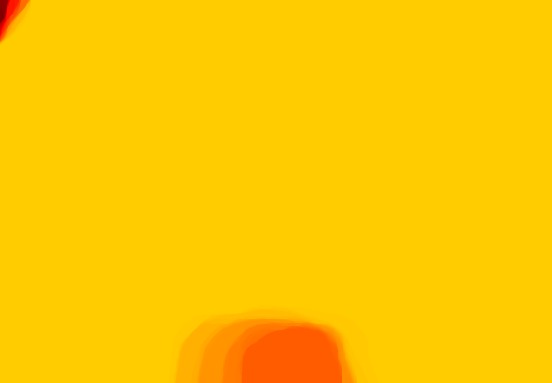}&
		\includegraphics[width=0.115\textwidth,trim={0cm 4cm 5cm 0}, clip]{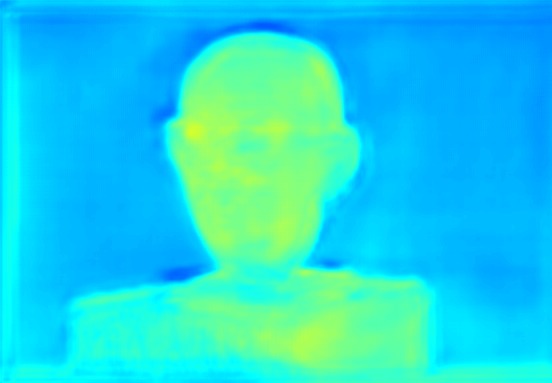}&
		\includegraphics[width=0.115\textwidth,trim={0cm 4cm 5cm 0}, clip]{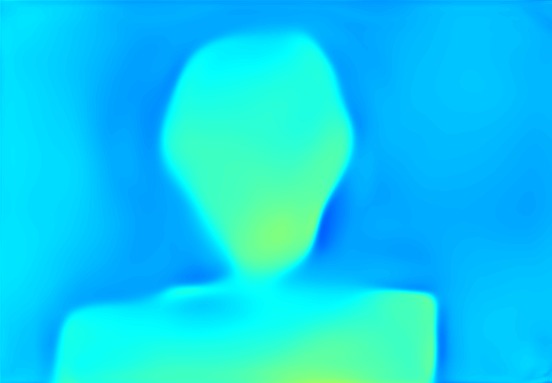}&	
		\includegraphics[width=0.115\textwidth,trim={0cm 4cm 5cm 0}, clip]{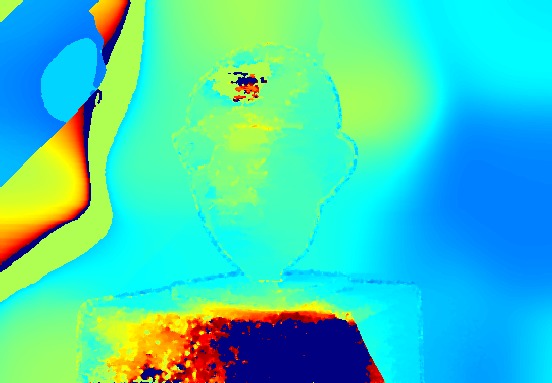}&
		\includegraphics[width=0.115\textwidth,trim={0cm 4cm 5cm 0}, clip]{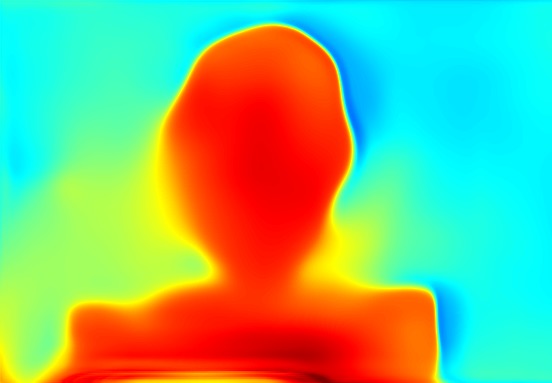}&		
		\includegraphics[width=0.115\textwidth,trim={0cm 4cm 5cm 0}, clip]{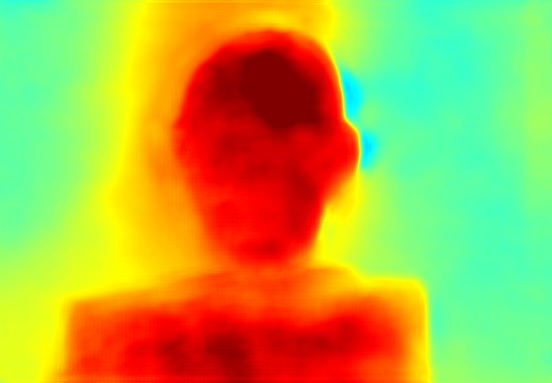}\\
		
		\includegraphics[width=0.115\textwidth,trim={0cm 4cm 5cm 0}, clip]{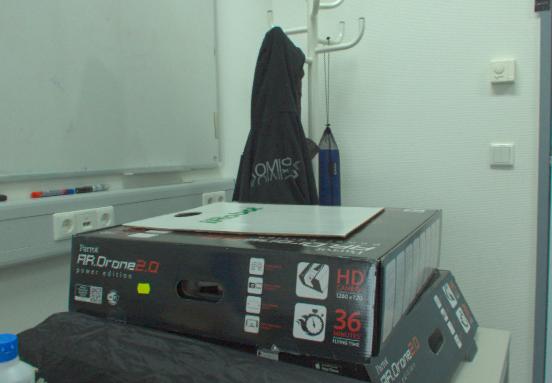}&
		\includegraphics[width=0.115\textwidth,trim={0cm 4cm 5cm 0}, clip]{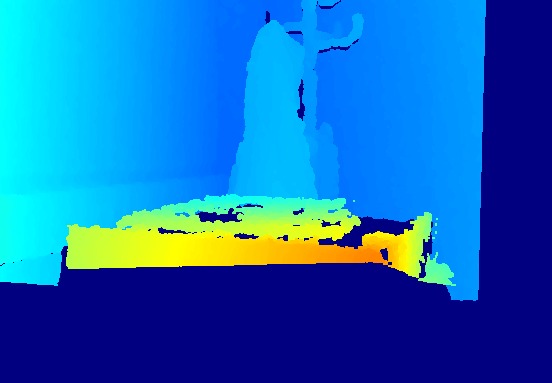}&
		\includegraphics[width=0.115\textwidth,trim={0cm 4cm 5cm 0}, clip]{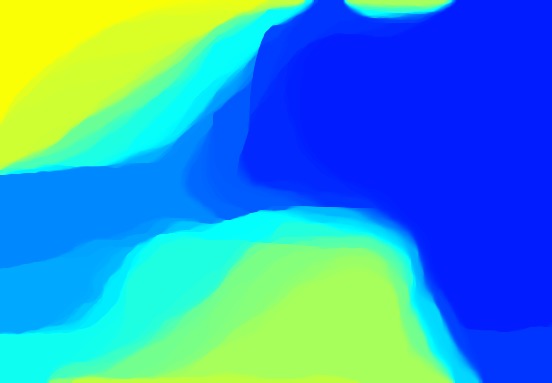}&
		\includegraphics[width=0.115\textwidth,trim={0cm 4cm 5cm 0}, clip]{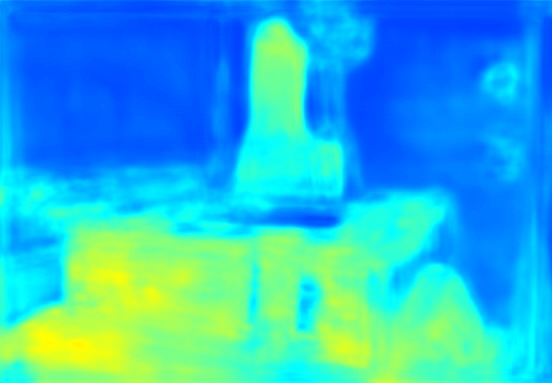}&
		\includegraphics[width=0.115\textwidth,trim={0cm 4cm 5cm 0}, clip]{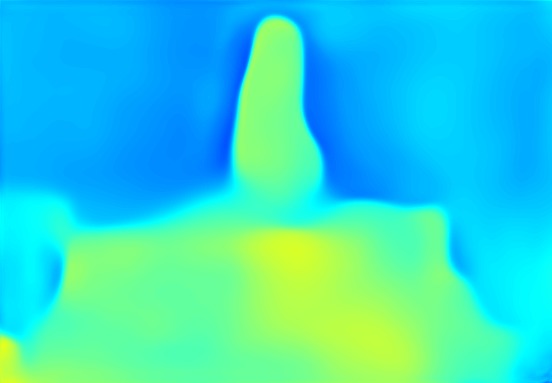}&
		\includegraphics[width=0.115\textwidth,trim={0cm 4cm 5cm 0}, clip]{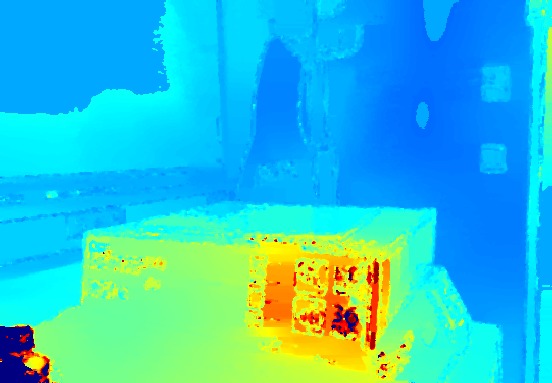}&
		\includegraphics[width=0.115\textwidth,trim={0cm 4cm 5cm 0}, clip]{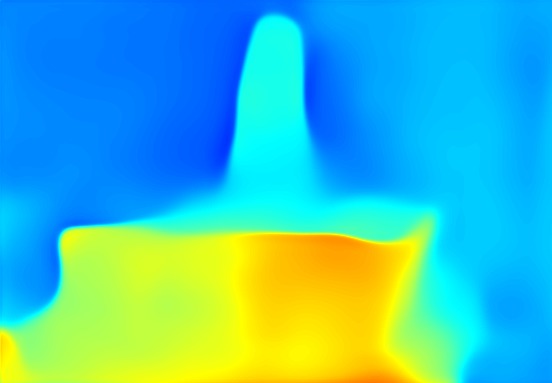}&
		\includegraphics[width=0.115\textwidth,trim={0cm 4cm 5cm 0}, clip]{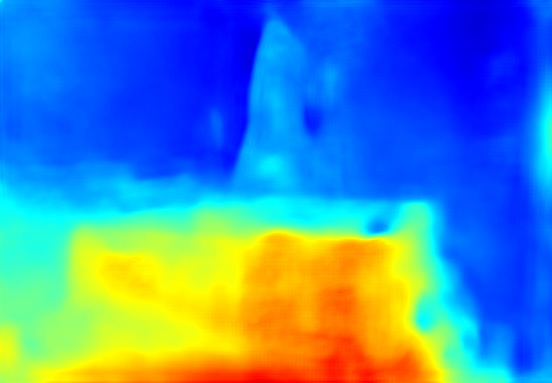}\\
		
		\includegraphics[width=0.115\textwidth,trim={0cm 4cm 5cm 0}, clip]{suppfigures/office44_IMG_0009}&
		\includegraphics[width=0.115\textwidth,trim={0cm 4cm 5cm 0}, clip]{suppfigures/office44_DISP_0009_gt}&
		\includegraphics[width=0.115\textwidth,trim={0cm 4cm 5cm 0}, 
		clip]{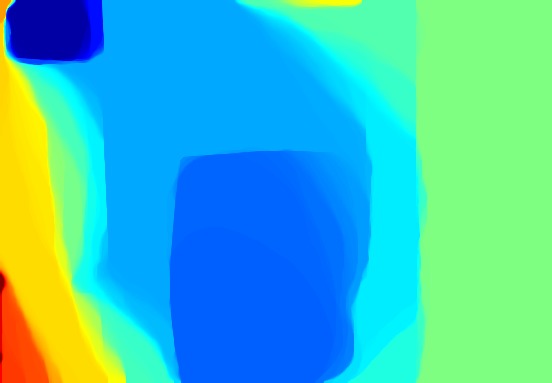}&
		\includegraphics[width=0.115\textwidth,trim={0cm 4cm 5cm 0}, 
		clip]{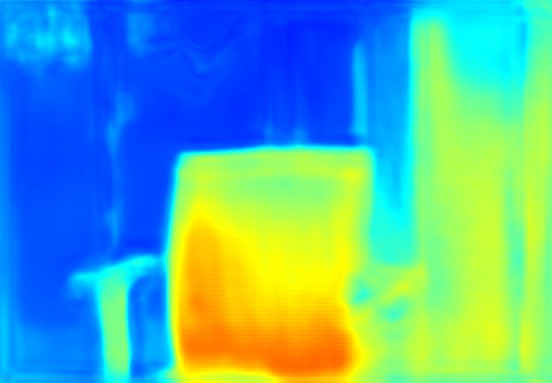}&
		\includegraphics[width=0.115\textwidth,trim={0cm 4cm 5cm 0}, 
		clip]{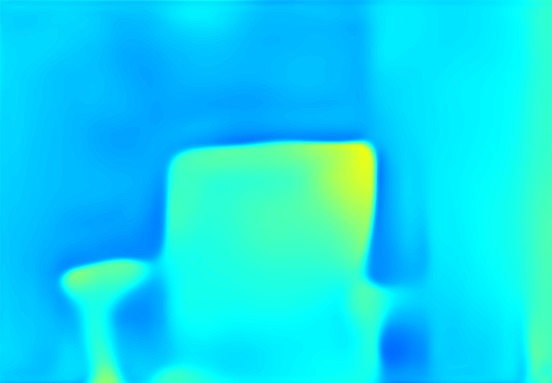}&
		\includegraphics[width=0.115\textwidth,trim={0cm 4cm 5cm 0}, clip]{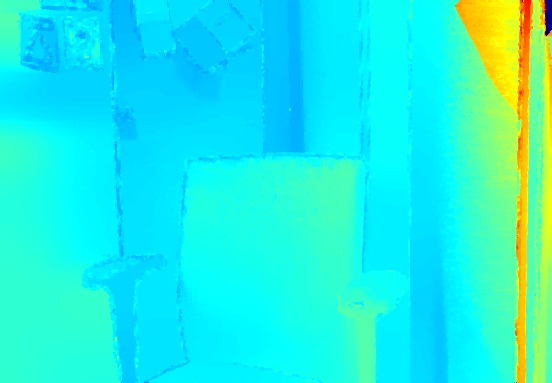}&
		\includegraphics[width=0.115\textwidth,trim={0cm 4cm 5cm 0}, 
		clip]{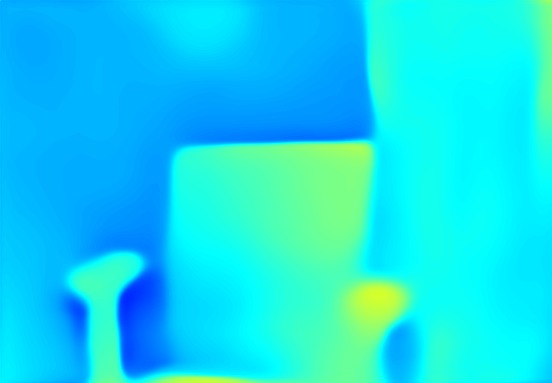}&
		\includegraphics[width=0.115\textwidth,trim={0cm 4cm 5cm 0}, clip]{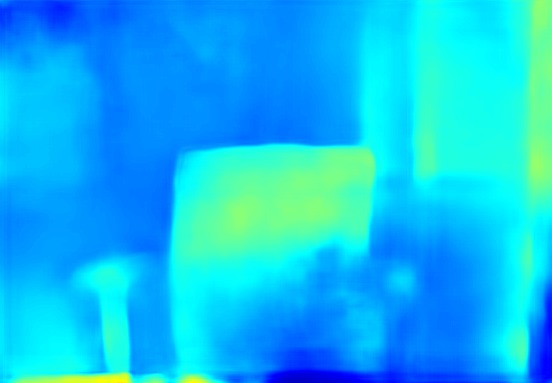}\\	
		
		\includegraphics[width=0.115\textwidth,trim={0cm 4cm 5cm 0}, clip]{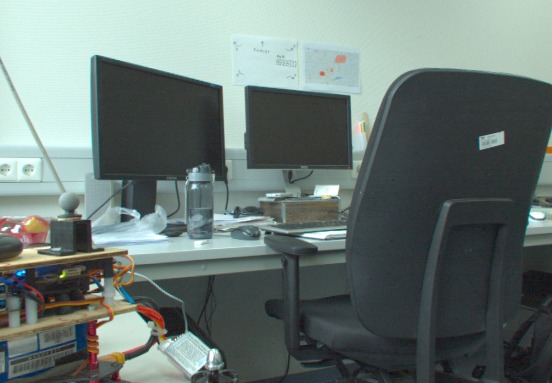}&
		\includegraphics[width=0.115\textwidth,trim={0cm 4cm 5cm 0}, clip]{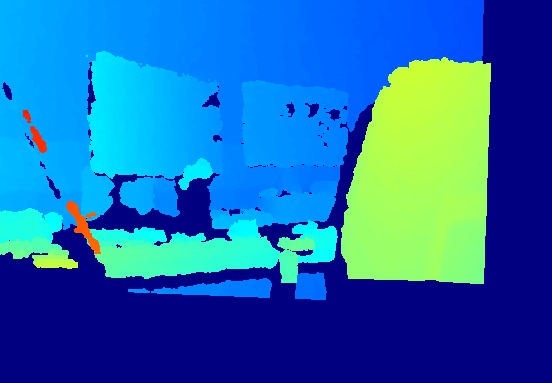}&
		\includegraphics[width=0.115\textwidth,trim={0cm 4cm 5cm 0}, clip]{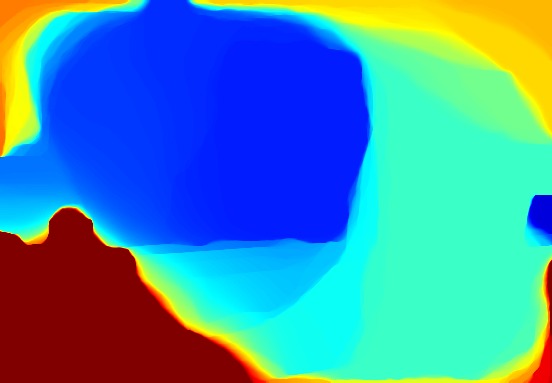}&
		\includegraphics[width=0.115\textwidth,trim={0cm 4cm 5cm 0}, clip]{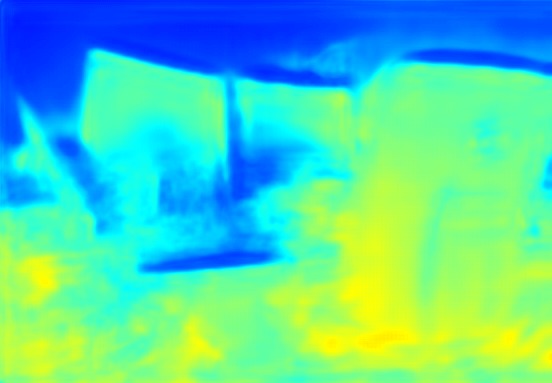}&
		\includegraphics[width=0.115\textwidth,trim={0cm 4cm 5cm 0}, clip]{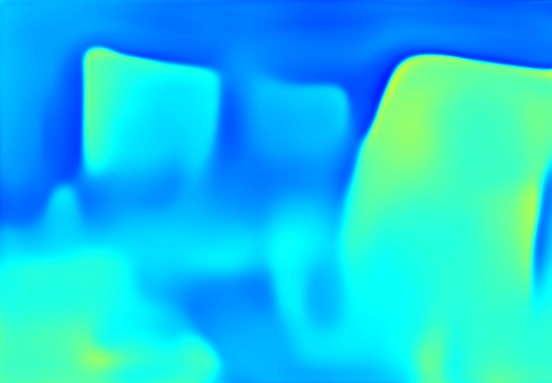}&			
		\includegraphics[width=0.115\textwidth,trim={0cm 4cm 5cm 0}, clip]{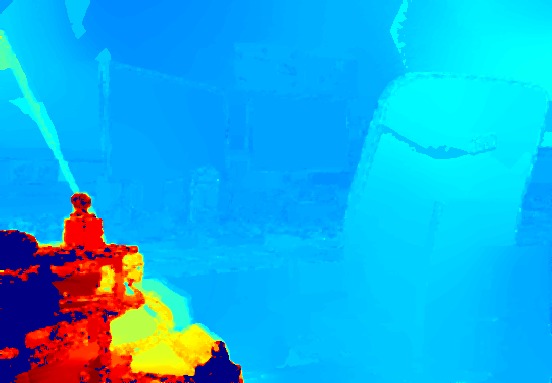}&
		\includegraphics[width=0.115\textwidth,trim={0cm 4cm 5cm 0}, clip]{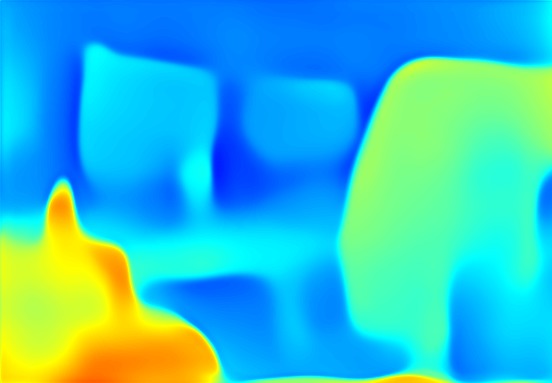}&
		\includegraphics[width=0.115\textwidth,trim={0cm 4cm 5cm 0}, clip]{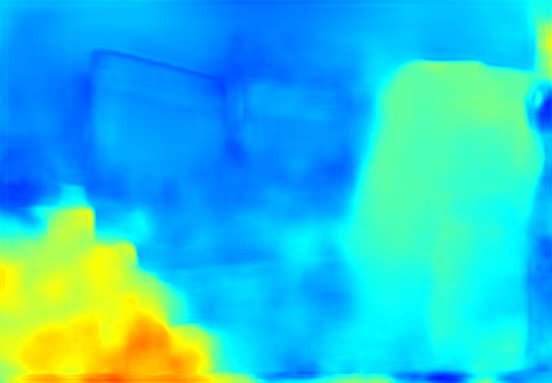}\\			
		
		\includegraphics[width=0.115\textwidth,trim={0cm 4cm 5cm 0}, clip]{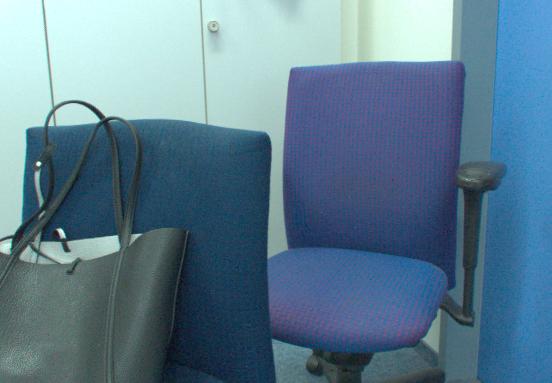}&
		\includegraphics[width=0.115\textwidth,trim={0cm 4cm 5cm 0}, clip]{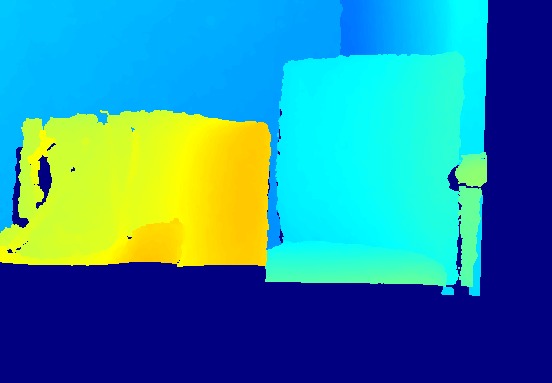}&
		\includegraphics[width=0.115\textwidth,trim={0cm 4cm 5cm 0}, clip]{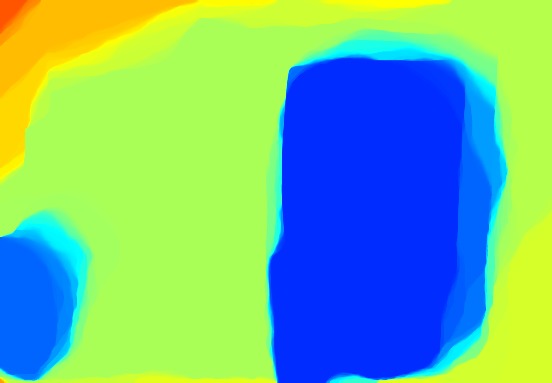}&
		\includegraphics[width=0.115\textwidth,trim={0cm 4cm 5cm 0}, clip]{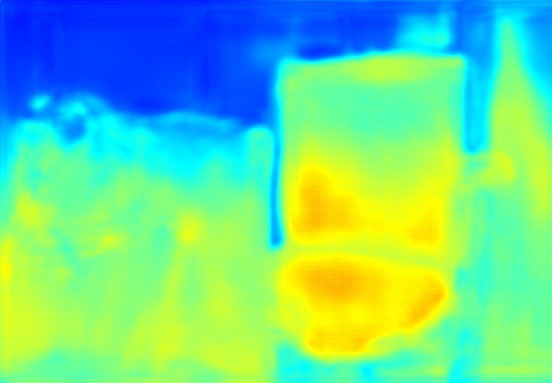}&
		\includegraphics[width=0.115\textwidth,trim={0cm 4cm 5cm 0}, clip]{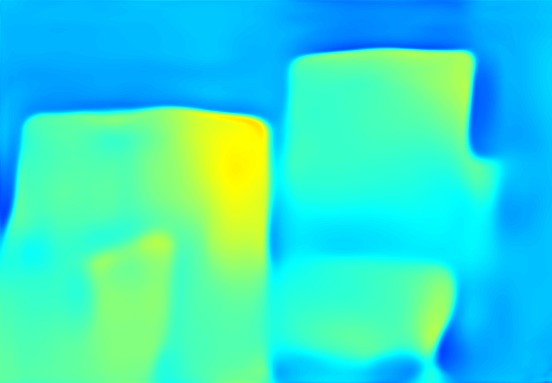}&
		\includegraphics[width=0.115\textwidth,trim={0cm 4cm 5cm 0}, clip]{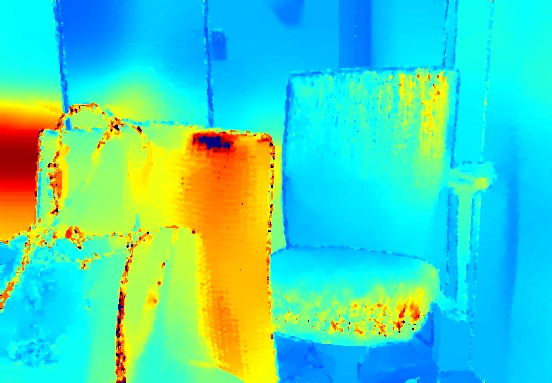}&
		\includegraphics[width=0.115\textwidth,trim={0cm 4cm 5cm 0}, clip]{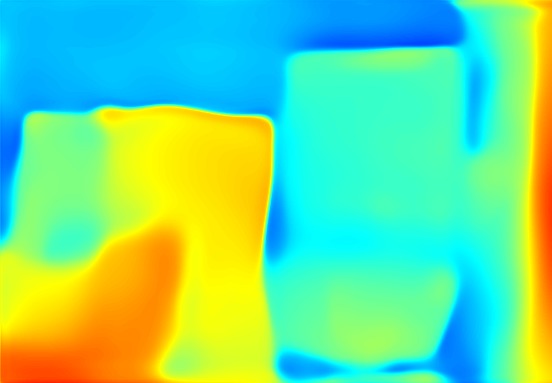}&
		\includegraphics[width=0.115\textwidth,trim={0cm 4cm 5cm 0}, clip]{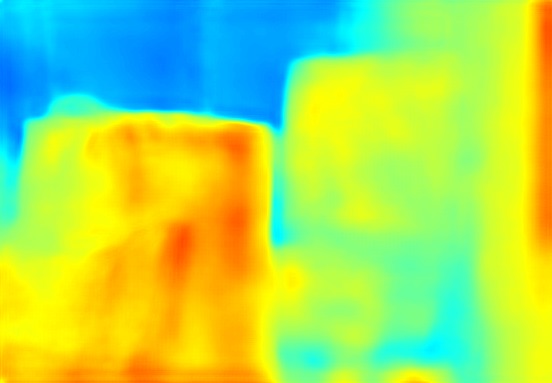}\\	
		
		\includegraphics[width=0.115\textwidth,trim={0cm 4cm 5cm 0}, clip]{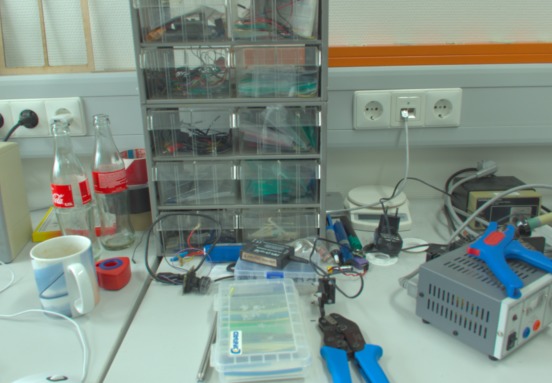}&
		\includegraphics[width=0.115\textwidth,trim={0cm 4cm 5cm 0}, clip]{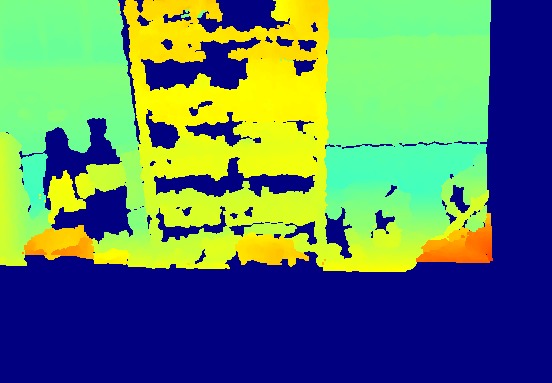}&
		\includegraphics[width=0.115\textwidth,trim={0cm 4cm 5cm 0}, clip]{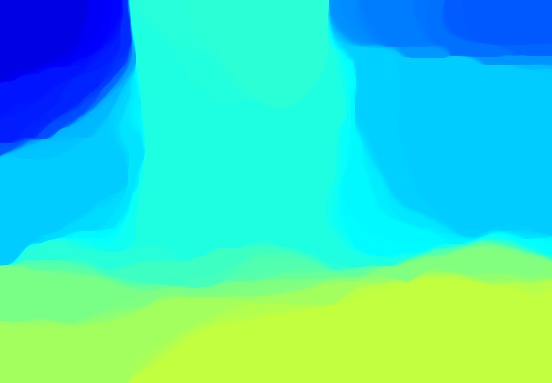}&
		\includegraphics[width=0.115\textwidth,trim={0cm 4cm 5cm 0}, clip]{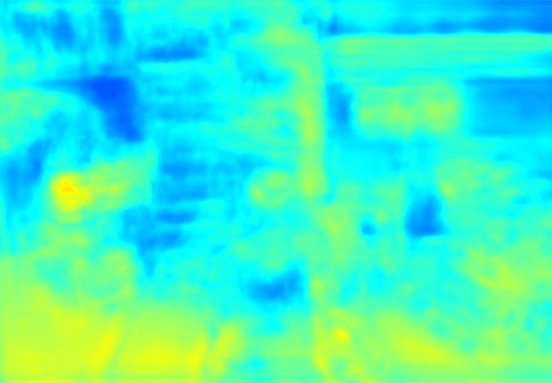}&
		\includegraphics[width=0.115\textwidth,trim={0cm 4cm 5cm 0}, clip]{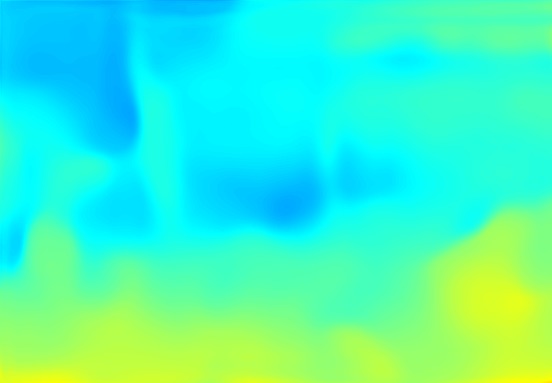}&
		\includegraphics[width=0.115\textwidth,trim={0cm 4cm 5cm 0}, clip]{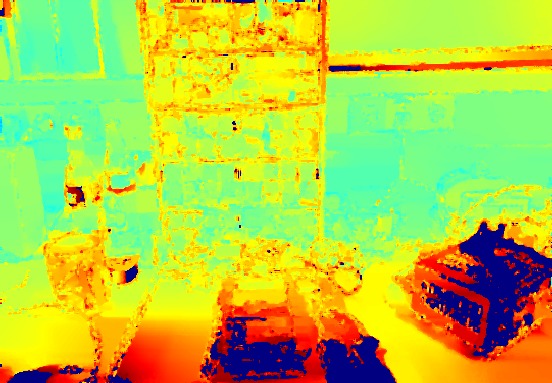}&
		\includegraphics[width=0.115\textwidth,trim={0cm 4cm 5cm 0}, clip]{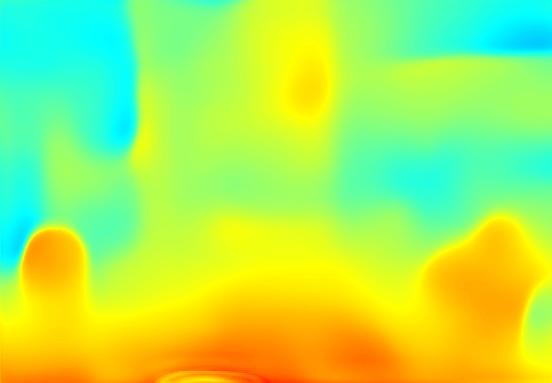}&
		\includegraphics[width=0.115\textwidth,trim={0cm 4cm 5cm 0}, clip]{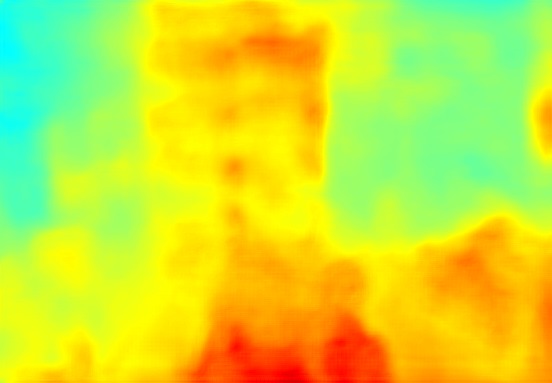}\\			
	\end{tabular}
	\caption{\small \textbf{Qualitative results of the DDFFNet versus state-of-the-art methods.} Results are normalized by the maximum disparity. Warmer colors represent closer distances. Best viewed in color.}
	\label{fig:qualRes}
\end{figure}

\begin{figure}
	\def\colsep{1pt}
	\def\rowstretch{0.5}
	\centering
	\begin{tabular}{c c c c c c c c}
		Image & Disparity & VDFF & DDLF & PSP-LF & Lytro & PSPNet & Proposed
		\\
		\includegraphics[width=0.115\textwidth,trim={0cm 4cm 5cm 0}, clip]{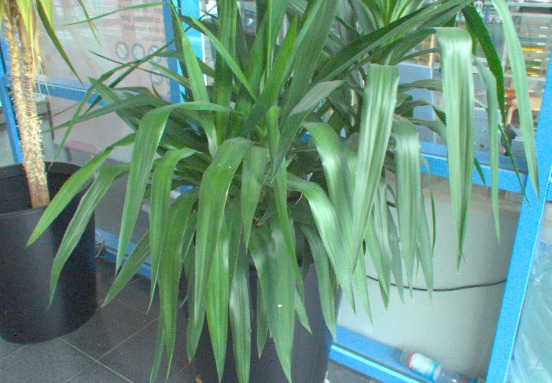}&
		\includegraphics[width=0.115\textwidth,trim={0cm 4cm 5cm 0}, clip]{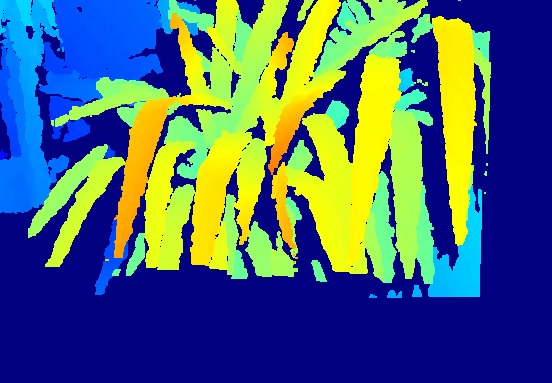}&
		\includegraphics[width=0.115\textwidth,trim={0cm 4cm 5cm 0}, clip]{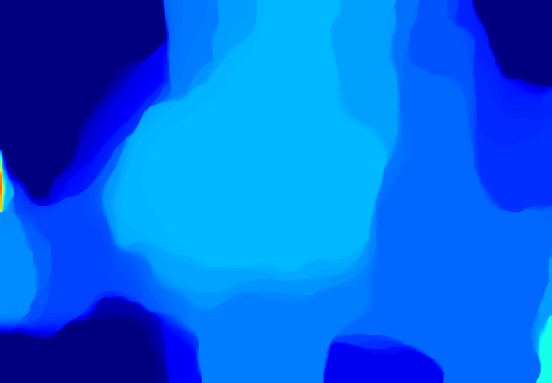}&
		\includegraphics[width=0.115\textwidth,trim={0cm 4cm 5cm 0}, clip]{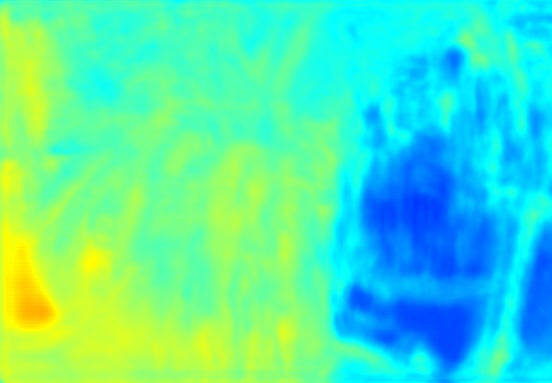}&
		\includegraphics[width=0.115\textwidth,trim={0cm 4cm 5cm 0}, clip]{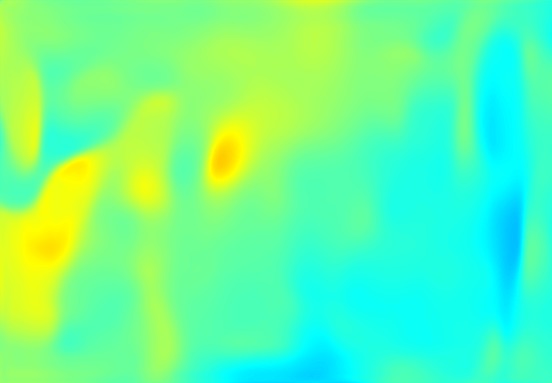}&
		\includegraphics[width=0.115\textwidth,trim={0cm 4cm 5cm 0}, clip]{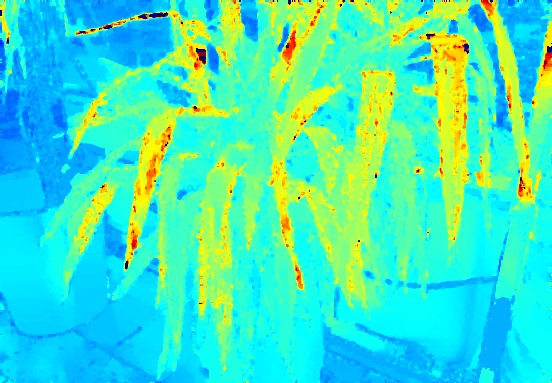}&
		\includegraphics[width=0.115\textwidth,trim={0cm 4cm 5cm 0}, clip]{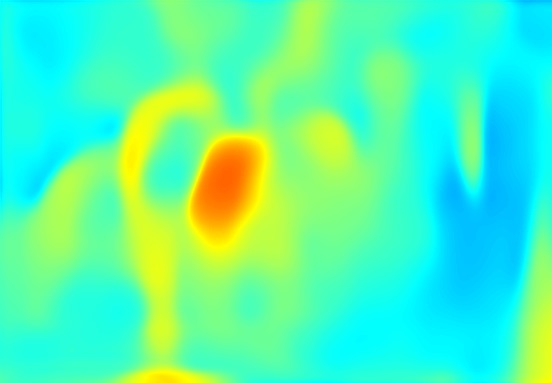}&
		\includegraphics[width=0.115\textwidth,trim={0cm 4cm 5cm 0}, clip]{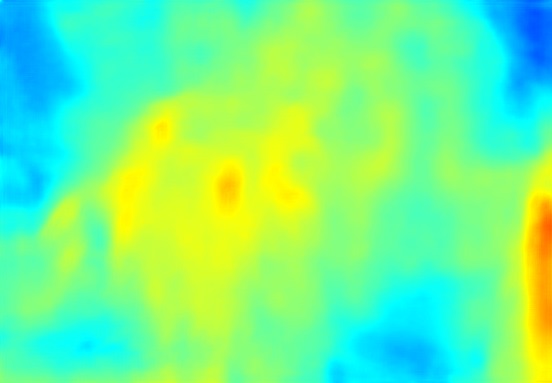}\\
		
		\includegraphics[width=0.115\textwidth,trim={0cm 4cm 5cm 0}, clip]{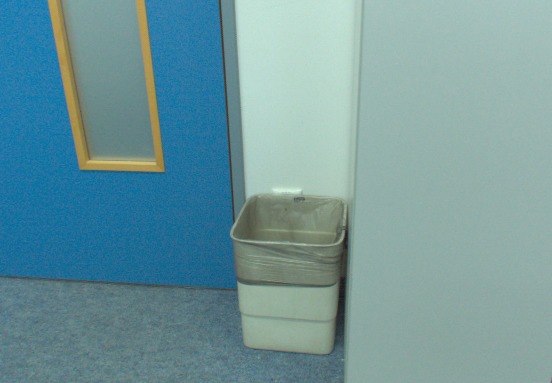}&
		\includegraphics[width=0.115\textwidth,trim={0cm 4cm 5cm 0}, clip]{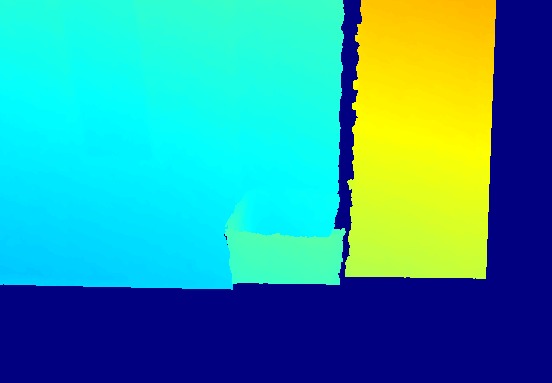}&
		\includegraphics[width=0.115\textwidth,trim={0cm 4cm 5cm 0}, clip]{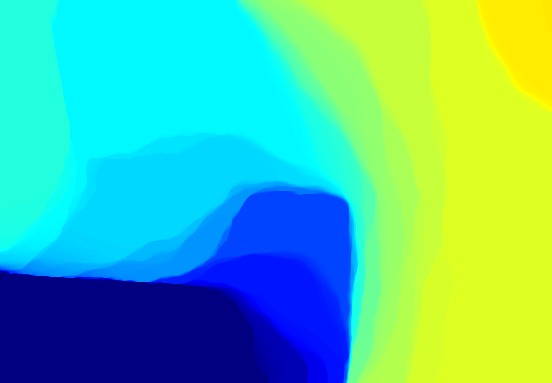}&
		\includegraphics[width=0.115\textwidth,trim={0cm 4cm 5cm 0}, clip]{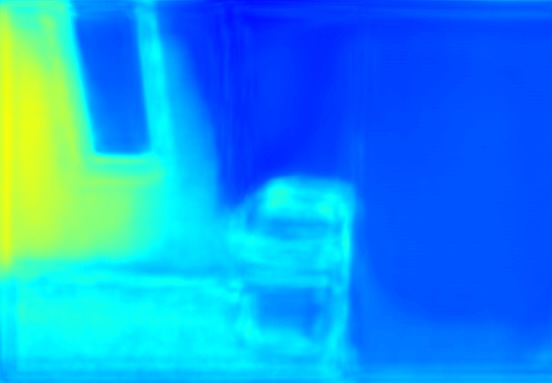}&
		\includegraphics[width=0.115\textwidth,trim={0cm 4cm 5cm 0}, clip]{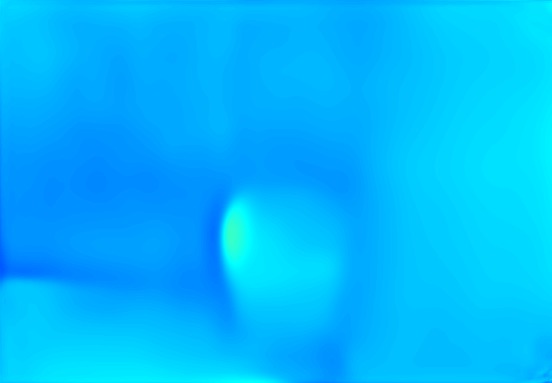}&			
		\includegraphics[width=0.115\textwidth,trim={0cm 4cm 5cm 0}, clip]{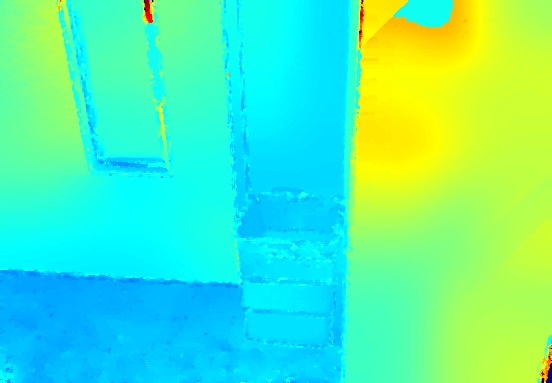}&
		\includegraphics[width=0.115\textwidth,trim={0cm 4cm 5cm 0}, clip]{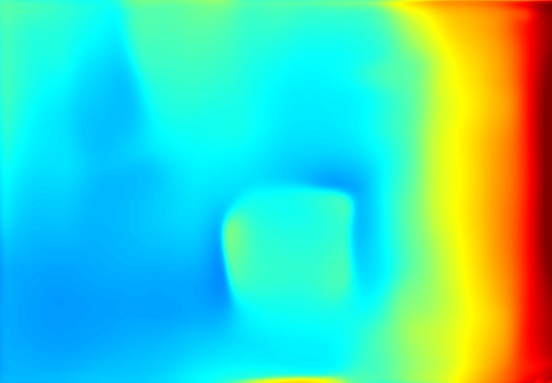}&
		\includegraphics[width=0.115\textwidth,trim={0cm 4cm 5cm 0}, clip]{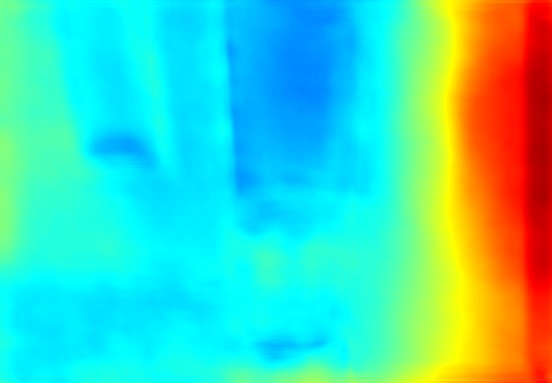}	
	\end{tabular}
	\caption{\small \textbf{Failure cases.} Results are normalized by the maximum disparity. Warmer colors represent closer distances. Best viewed in color.}
	\label{fig:qualResFail}
\end{figure}

\begin{figure}
	\def\colsep{1pt}
	\def\rowstretch{0.5}
	\centering
	\begin{tabular}{ccccc}
		Image & Disparity & VDFF & PSPNet & Proposed\\
		\includegraphics[width=0.192\textwidth]{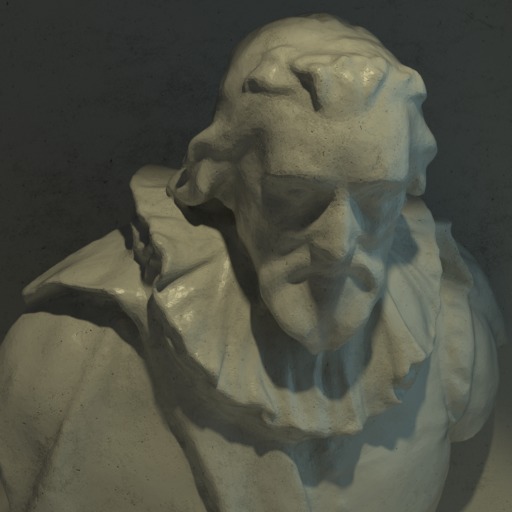} &
		\includegraphics[width=0.192\textwidth]{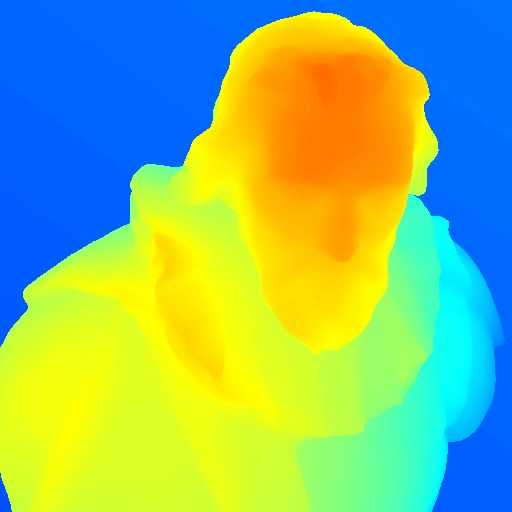} &
		\includegraphics[width=0.192\textwidth]{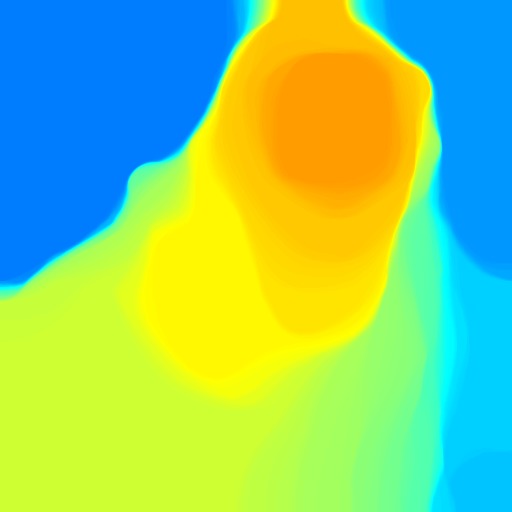} &
		\includegraphics[width=0.192\textwidth]{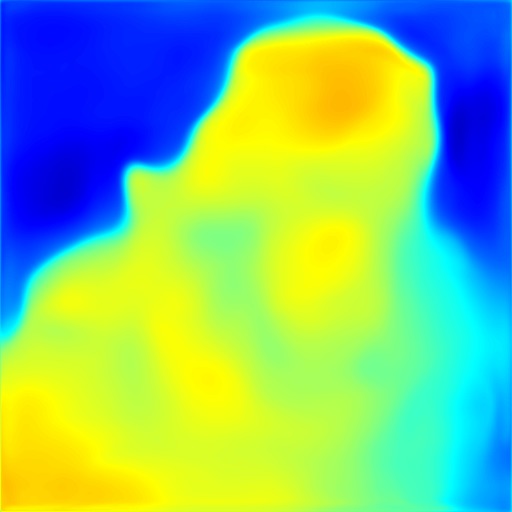} &
		\includegraphics[width=0.192\textwidth]{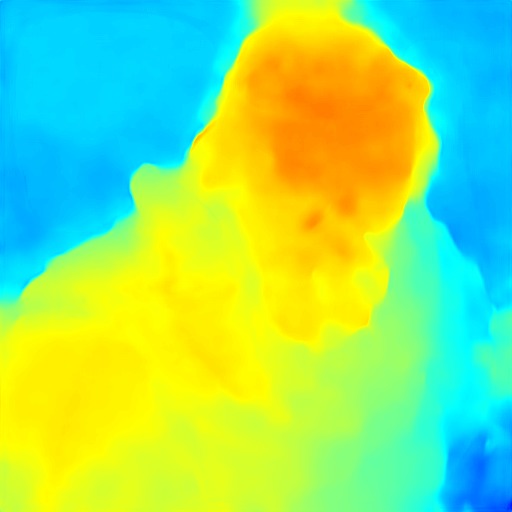}
		\\
		\includegraphics[width=0.192\textwidth]{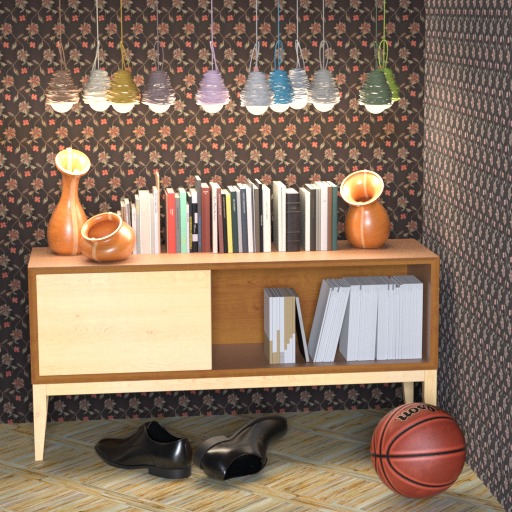} &
		\includegraphics[width=0.192\textwidth]{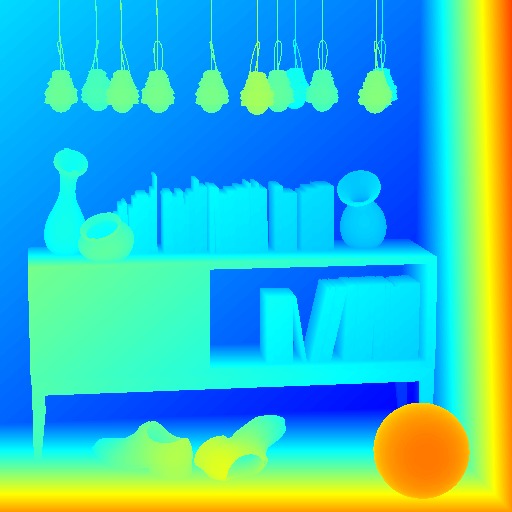} &
		\includegraphics[width=0.192\textwidth]{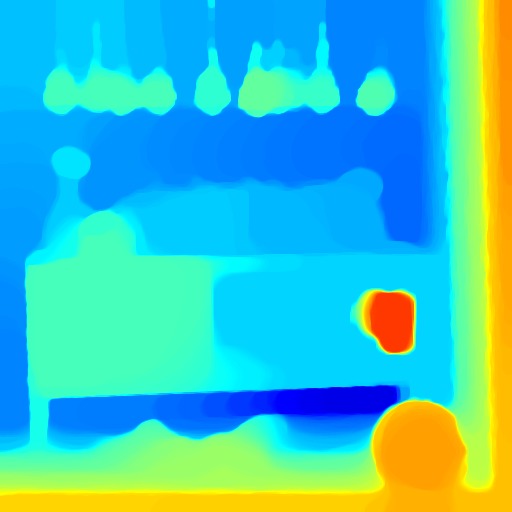} &			\includegraphics[width=0.192\textwidth]{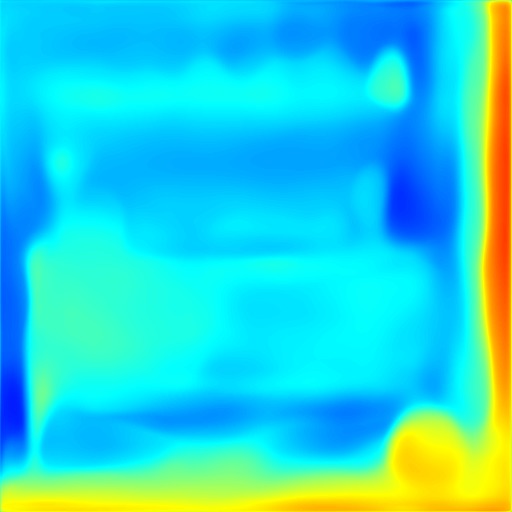} &			
		\includegraphics[width=0.192\textwidth]{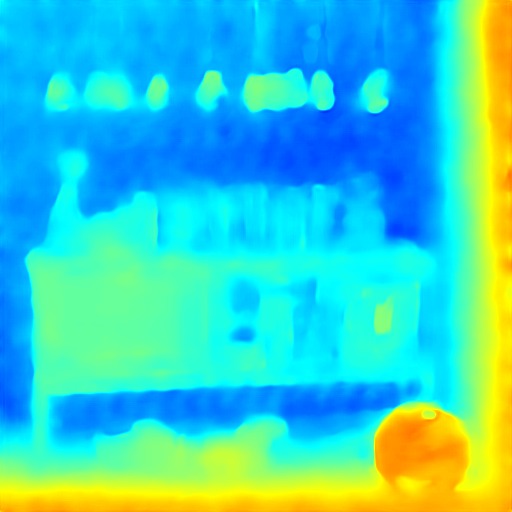}
		\\
		\includegraphics[width=0.192\textwidth]{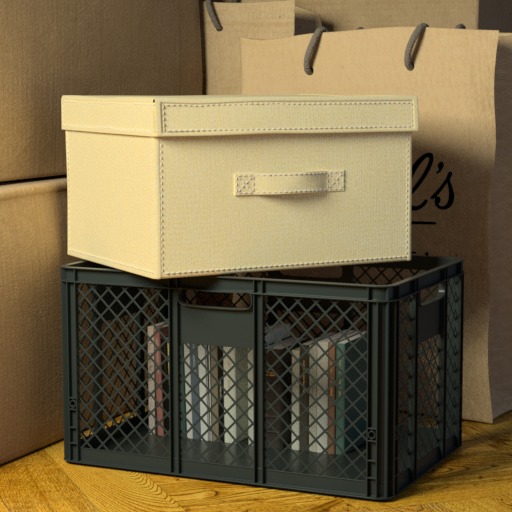} &
		\includegraphics[width=0.192\textwidth]{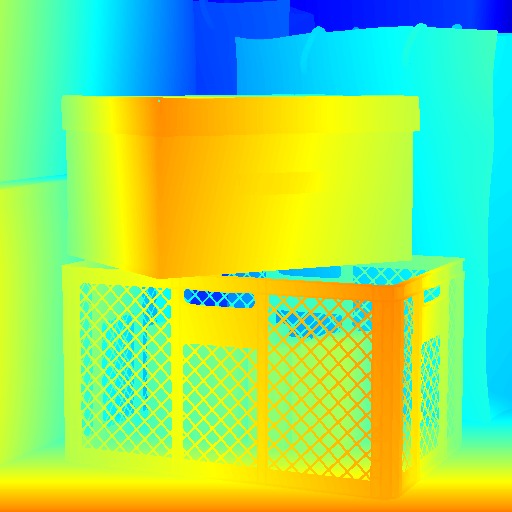} &
		\includegraphics[width=0.192\textwidth]{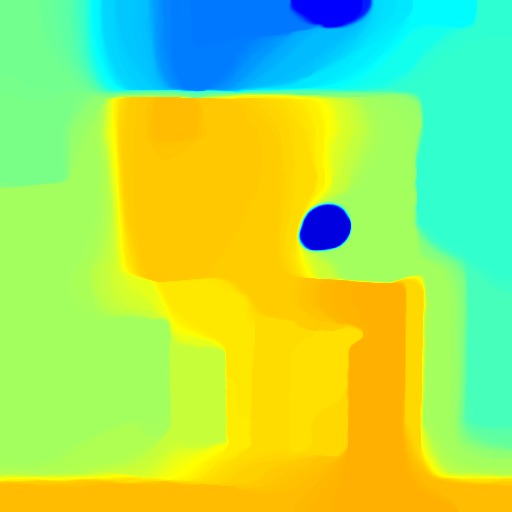}&
		\includegraphics[width=0.192\textwidth]{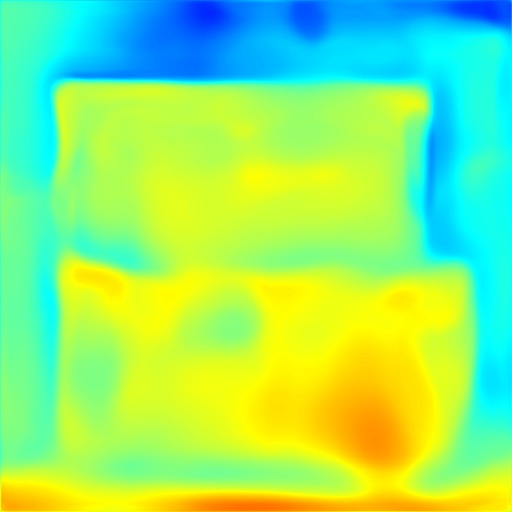}&
		\includegraphics[width=0.192\textwidth]{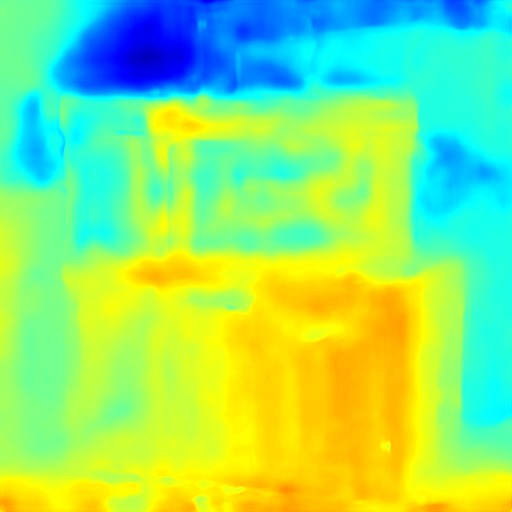}
		\\
		\includegraphics[width=0.192\textwidth]{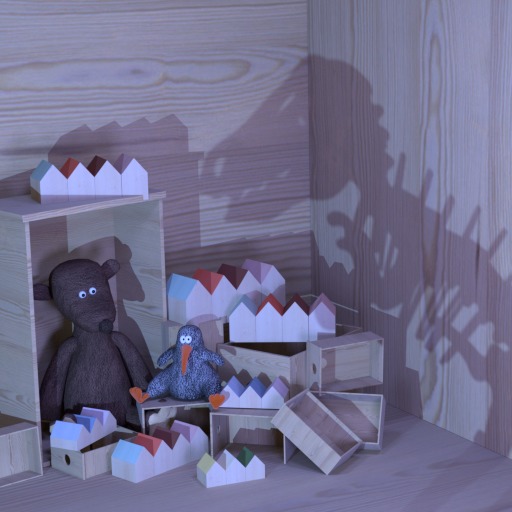} &
		\includegraphics[width=0.192\textwidth]{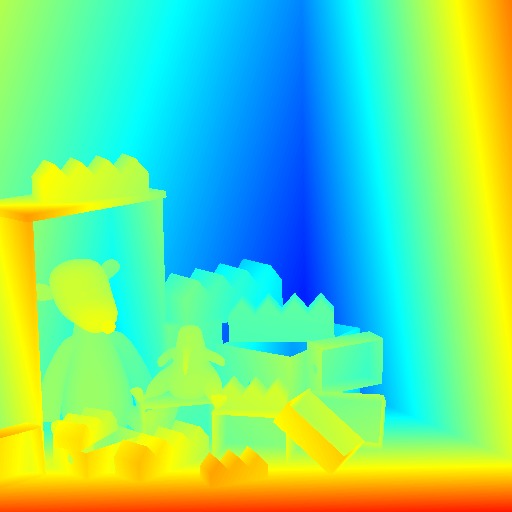} &
		\includegraphics[width=0.192\textwidth]{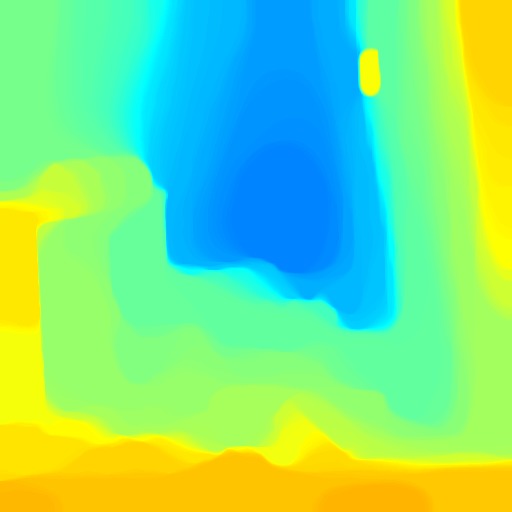}&
		\includegraphics[width=0.192\textwidth]{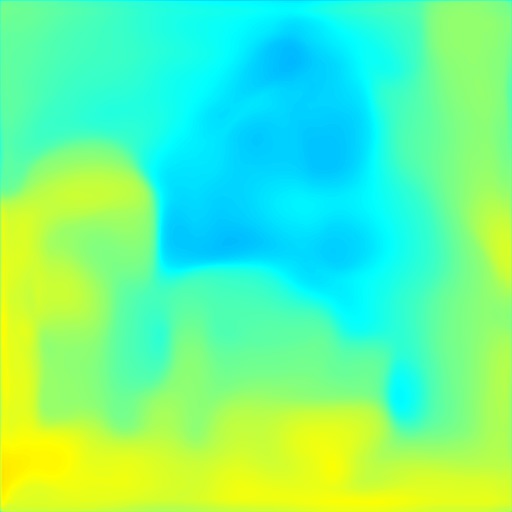}&
		\includegraphics[width=0.192\textwidth]{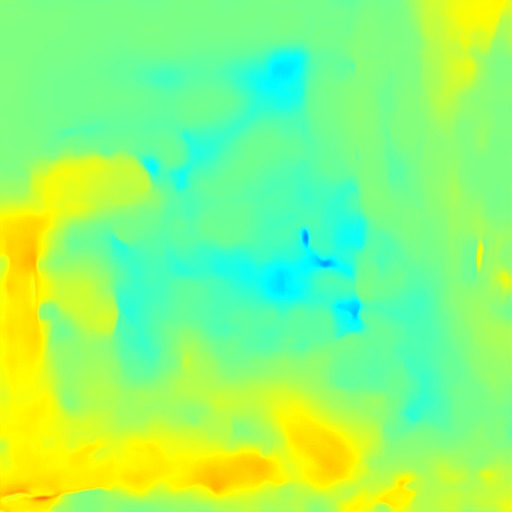}
	\end{tabular}
	\caption{\small \textbf{Results on the 4D light-field benchmark}~\cite{honauer16benchmark}. Our method preserves sharper object boundaries while other methods over-smooth the details in the estimated disparity maps.}
	\label{fig:4dlfdataset}
\end{figure}

\begin{figure}
	\def\colsep{1pt}
	\def\rowstretch{0.5}
	\centering
	\begin{tabular}{ccccc}
		Image & Depth & VDFF & PSPNet & Proposed\\
		\includegraphics[width=0.192\textwidth]{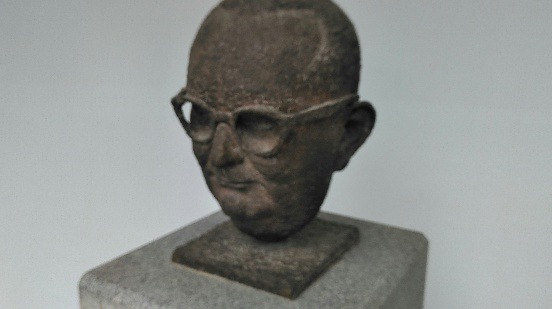}&
		\includegraphics[width=0.192\textwidth]{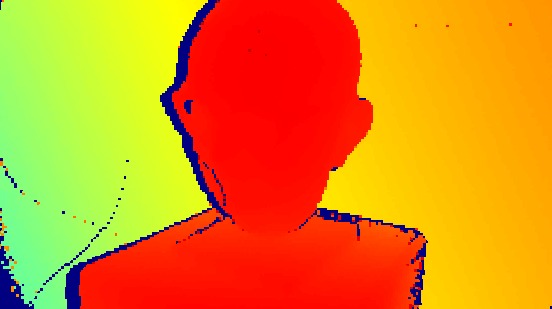}&
		\includegraphics[width=0.192\textwidth]{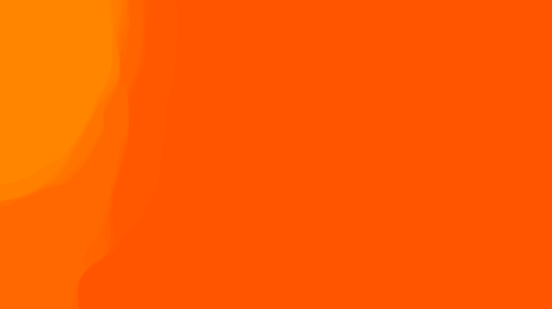}&
		\includegraphics[width=0.192\textwidth]{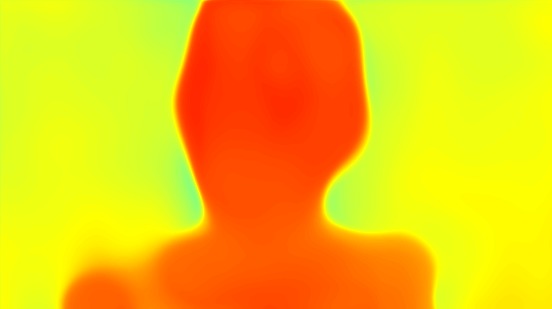}&
		\includegraphics[width=0.192\textwidth]{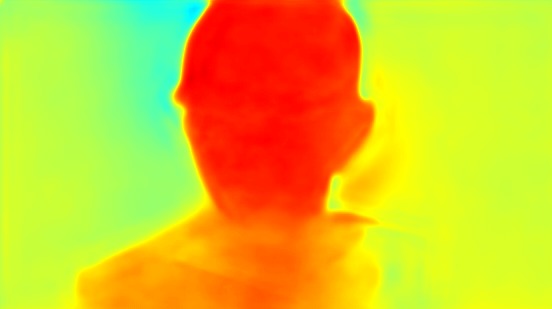}
		\\
		\includegraphics[width=0.192\textwidth]{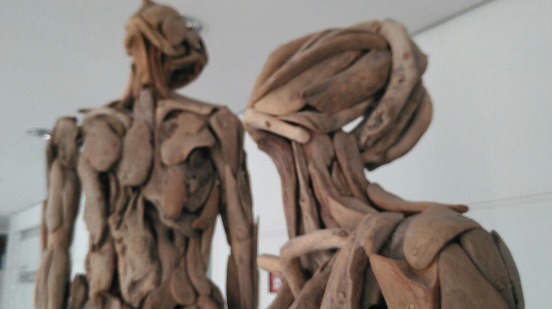}&
		\includegraphics[width=0.192\textwidth]{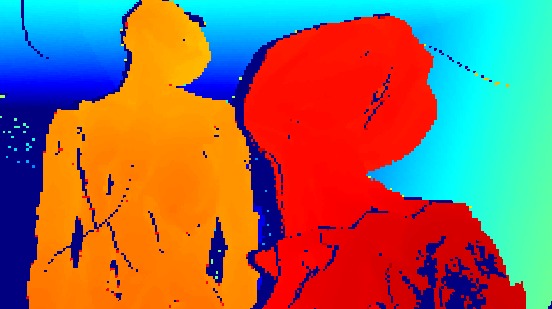}&
		\includegraphics[width=0.192\textwidth]{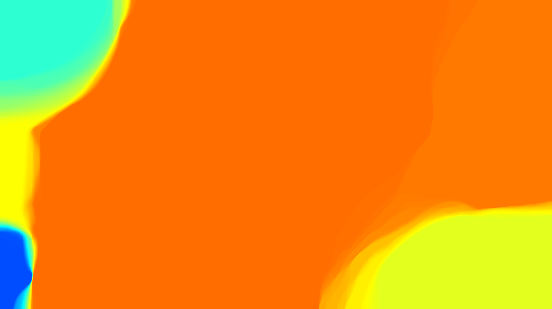}&		
		\includegraphics[width=0.192\textwidth]{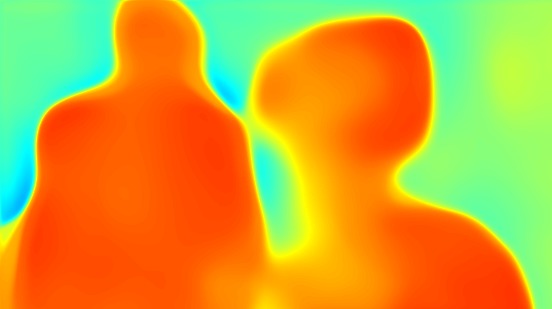}&
		\includegraphics[width=0.192\textwidth]{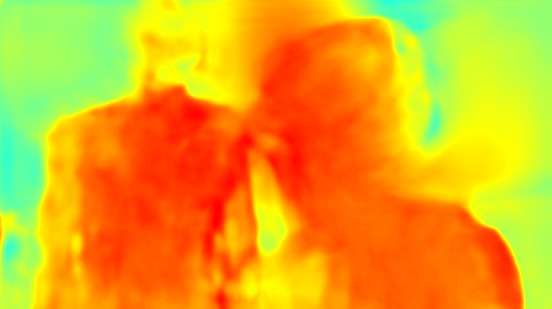}
	\end{tabular}
	\caption{\label{fig:android}\textbf{Results on the mDFF dataset.} PSPNet produces smooth boundaries while \textit{DDFFNet-CC3} keeps sharp edges in the results.}
\end{figure}

\end{document}